\documentclass[10pt,twocolumn,letterpaper]{article}

\usepackage{times}
\usepackage{latexsym}
\usepackage{float}
\usepackage{xcolor}
\usepackage{amsmath}
\usepackage{amssymb}
\usepackage{subcaption}
\usepackage{fvextra}
\usepackage{framed}
\usepackage{tcolorbox}
\tcbuselibrary{breakable, skins}
\usepackage{multirow}
\usepackage{makecell}
\usepackage{colortbl}
\usepackage{dblfloatfix}
\usepackage[pagenumbers]{wacv} 
\definecolor{wacvblue}{rgb}{0.21,0.49,0.74}
\usepackage[pagebackref,breaklinks,colorlinks,allcolors=wacvblue]{hyperref}

\def\wacvPaperID{*****} 
\def\confName{WACV}
\def\confYear{2027}

\title{Chehre: An Emoji-Prompted Dataset to Explore \\ Perceptual Flexibility in Video Language Models}

\author{
Bita Azari, Zoe Stanley, Avneet Batra, Poorvi Bhatia,\\
Hali Kil, Manolis Savva, Angelica Lim\\[0.3em]
Simon Fraser University, Canada\\[0.3em]
\textbf{Correspondence:} \texttt{bazari@sfu.ca}
}

\begin{document}
\maketitle
\begin{abstract}
Do people perceive the same facial expression in the same way? Should we expect vision models to be flexible in how they perceive facial expressions? Facial expressions are nonverbal social signals used in human interaction, but facial expression recognition datasets often focus on a single deterministic annotation per sample. We introduce \textit{Chehre}, an emoji-prompted video dataset with a wide range of dynamic facial expressions for exploring perceptual variation. In \textit{Chehre}, 203 participants were prompted to express and record 40 facial emojis. Later, their facial motions were transferred onto synthetic faces to preserve privacy. A separate group annotated the videos, resulting in 2,111 videos annotated by 1,242 perceivers, with $\sim$30 annotators per video. Chehre enables us to define a new task: ``distributional expression recognition", which tests whether a model can reproduce the variation observed across annotator responses. We test a selection of video language models on our task. Interestingly, we find that persona prompting can act as a controllable way to shift model perception while helping models better capture the variation observed across human annotators.
The dataset and code are available at \url{https://chehre-dataset.github.io/.}


\end{abstract}
    

\begin{figure}[!t]
    \centering

    \includegraphics[width=\columnwidth]{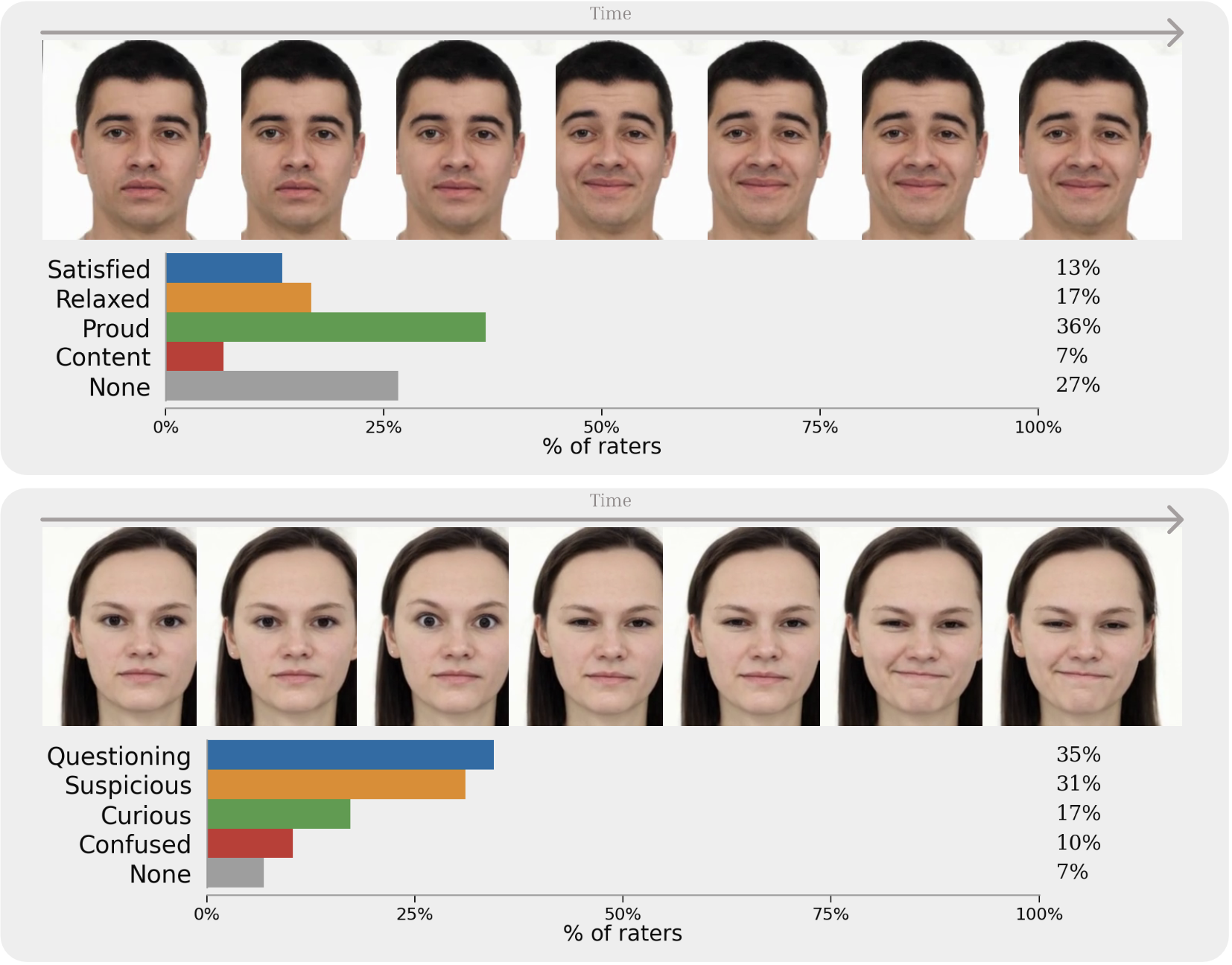}
    \caption{Anonymized and annotated videos from \textit{Chehre}. The bar plot shows the percentage of annotators who selected each label as top-1 from the candidate set of labels for each video.}
    \label{fig:first-page-photo}

    \vspace{0.8em}

    \includegraphics[width=\columnwidth]{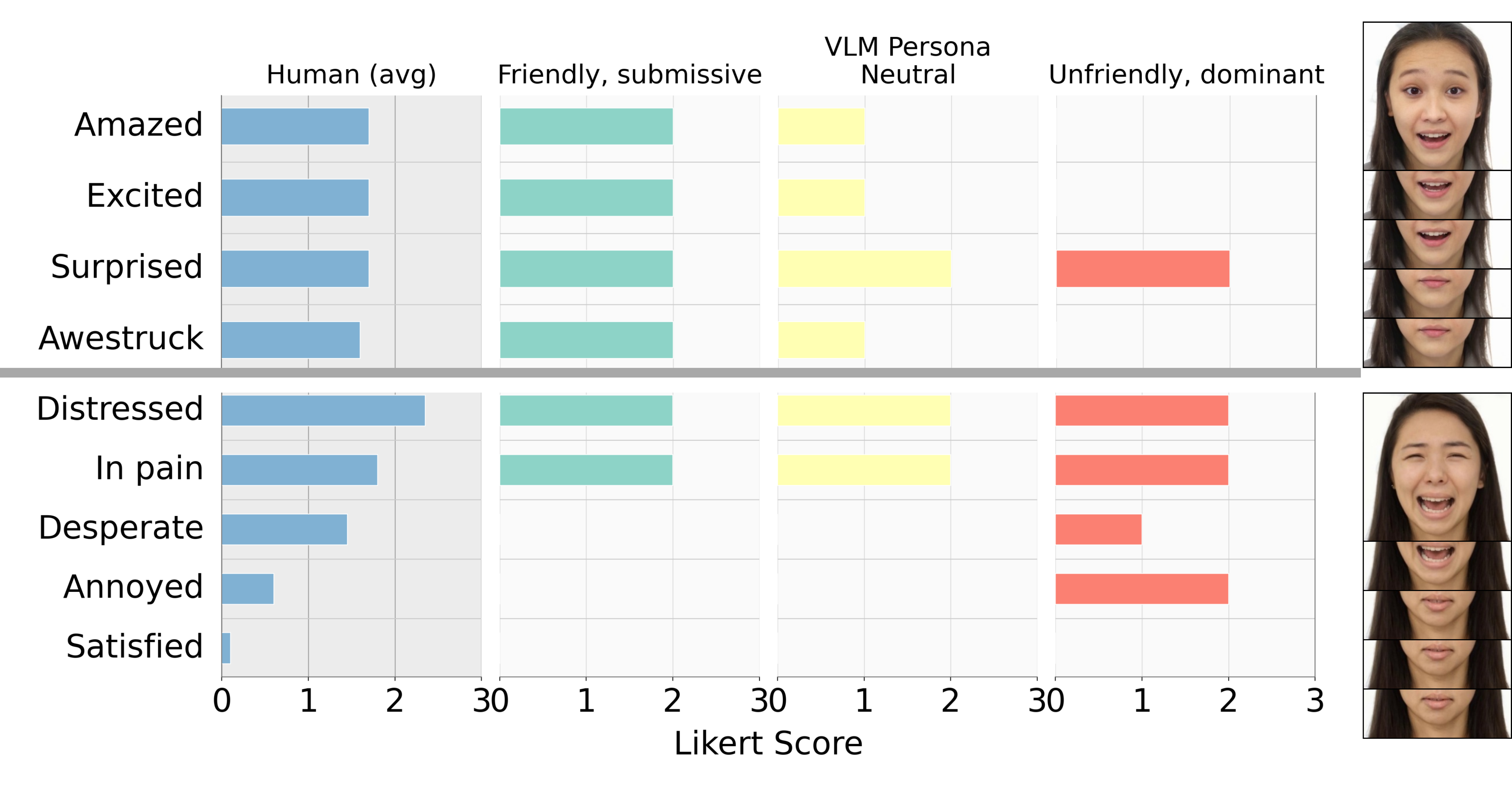}
    \caption{Two sample videos analyzed with Qwen2.5-Omni, showing the effect of persona prompting. Top: With an unfriendly and dominant persona, positive emotions such as amazed and excited are not output, leaving only surprised. Bottom: Prompted to be unfriendly and dominant, the VLM perceives the expression as more desperate and annoyed.}
    \label{fig:persona}
\end{figure}
\section{Introduction}
As embodied AI agents begin to interact naturally with humans, multimodal language models are tasked with interpreting \textit{nonverbal} social signals, such as facial expressions and body language \cite{etesam2024contextual}. Yet, facial expression recognition (or ``emotion recognition"\footnote{While ``emotion recognition" benchmarks are common, we avoid this term to expand the set of terms in the dataset. Facial expressions also do not necessarily reflect a person's internal feeling.}) remains an extremely challenging task, because annotators rarely agree on a single label for an expression \cite{barrett2019emotional}. Benchmarks currently simplify the task, assuming a single definitive label per facial expression based on majority responses. Many also focus on static images rather than dynamic videos representative of real-life interactions, or on a set of basic emotion categories. Overall, the community lacks datasets that  capture the large variation in nonverbal social signals used in human communication, along with the variation in perception observed among individuals.



Emojis are a widely used way of expressing nonverbal social signals beyond emotions, from attitudes (e.g. affection, sarcasm, confidence), bodily states (e.g. tiredness, dizziness); and mental states (e.g. thinking, confusion). Therefore, they can provide practical elicitation prompts for studying a broader range of facial signals. Although emojis are familiar, their meanings are not fixed, allowing participants to interpret and enact them in different ways.
In this work, we introduce \textit{Chehre}, an emoji-prompted video dataset (Figure~\ref{fig:first-page-photo}). Participants were asked to express 40 emojis using face and head movements. We used facial reenactment to anonymize their videos and asked a separate group of annotators to evaluate the performances. Chehre enables us to define a new task: “distributional expression recognition”, which tests whether a model can reproduce the variation across human responses.

In summary, our contributions are:
\begin{itemize}
    \setlength{\itemsep}{0.25em}
    \setlength{\parskip}{0pt}
    \setlength{\parsep}{0pt}
    \setlength{\topsep}{0.2em}
    \item We introduce \textit{Chehre}, a controlled and anonymized emoji-prompted dataset of 2,111 dynamic synthetic-face videos that captures a large variety of social signals, collected and annotated with nearly $1500$ participants and ~30 annotators per video.
    \item Using Chehre, we explore perceptual flexibility in video language models by testing recent VLMs on a ``distributional recognition" task to reproduce perception variation among human annotators.
    \item To the best of our knowledge, we are the first to test if persona prompting affects how VLMs perceive facial expressions.
\end{itemize}
\section{Related Work}
\textbf{Facial Expression Datasets} Early facial expression recognition methods focused on static images \cite{mollahosseini2017affectnet, li2017reliable, lyons1998japanese}. These datasets opened the possibility of studying and analyzing facial expressions, but they do not fully capture the full range of facial expressions as a social signal \cite{ambadar2005deciphering}. In recent years, more dynamic facial expression datasets have been introduced, such as MELD~\cite{poria2019meld}, DFEW \cite{jiang2020dfew}, CAER~\cite{lee2019context}, which use basic emotion categories, and Aff-Wild2  \cite{kollias2018aff} and VEATIC \cite{ren2024veatic}, which use continuous valence/arousal scales for annotations. In order to achieve scale, many video datasets are collected from movies or in-the-wild media, under uncontrolled conditions. 

\textbf{The need for additional affective categories} In psychology, basic emotion categories have been challenged as incomplete descriptions of facial expression \cite{barrett2019emotional}. Cowen et al. \cite{cowen2020face} also highlight the limitations of basic emotion categories and continuous valence/arousal representations. They showed that naturalistic facial expressions can convey 28 distinct affective dimensions mapped to smooth semantic gradients rather than a few categories. While their dataset is abundant, it contains uncontrolled visual conditions, and to our knowledge no annotated facial emotion dataset anonymizes participant identities. Chehre builds on this work by anonymizing and standardizing videos through facial reenactment, expanding the expression space to include social and bodily-state signals via emojis, and collecting more than 30 annotations per video. 

\textbf{Emojis as prompts} Emojis represent affective and social meaning. Prior work investigates emojis as quasi-nonverbal cues in digital communication, showing that they convey emotional intensity and valence and can elicit affective responses similar to facial expressions in face-to-face interaction \cite{erle2022emojis}. EmojiHeroVR~\cite{ortmann2024emojiherovr} uses emojis to prompt participants in a virtual reality setting, but faces are occluded by the head-mounted VR display. To our knowledge, no existing dataset collects full, unoccluded facial expression video recordings using emojis as prompts. Chehre uses emojis to elicit a wide variety of dynamic facial expressions, including attitudes, bodily states, and mental states, while keeping the full face visible throughout.

\textbf{Affective intelligence in language models} Recent work has evaluated large-scale models for affective understanding. Lian et al.~\cite{lian2024gpt} evaluate GPT-4V as a zero-shot model for generalized emotion recognition across visual, temporal, textual and multimodal affective tasks. Emotion-LLaMA~\cite{cheng2024emotion} studied multimodal emotion recognition and reasoning using different modalities, and EmoBench~\cite{sabour2024emobench} evaluated emotional intelligence in LLMs. Also, VIBE~\cite{chakraborty-etal-2025-vibe} and MME-Emotion~\cite{zhang2025mme} assess VLMs on social-pragmatic inference and emotional understanding from visual scenes. However, evaluating VLMs on isolated facial and head-motion videos with per-video candidate label sets and Likert-scale ratings remains an open problem. Chehre provides this evaluation setting, enabling a controlled comparison between model and human perception on dynamic, single-face videos.

\textbf{Facial expression perception is subjective} Facial expression recognition is a subjective task, and individuals can interpret faces differently~\cite{barrett2019emotional, mohanty2025top}. Perceptual variation has been attributed to cultural differences \cite{fang2021cultural, jack2012facial}, yet studies \cite{binetti2022genetic} also show substantial inter-individual variation exists even when controlling for culture. In recent years, studies in natural language processing and computer vision \cite{uma2021learning, plank2022problem, davani2021dealingdisagreementslookingmajority,geng2016label} have challenged the assumption that in many perception tasks there is only one correct single (gold) interpretation. Instead, they argue that disagreement should not be treated as noise but as a meaningful signal, a stance often referred to as the perspectivist approach. Recent work \cite{frohling2025personas, lutz2025prompt} has explored persona prompting to guide LLM responses and induce demographic variation. Chehre embraces the perspectivist approach through distributional annotation and is the first to investigate how persona prompting affects VLM-based expression recognition.
\section{Dataset}
The \textit{Chehre} dataset was constructed in two phases. In the \textbf{Expression Phase}, human performers were recorded expressing 40 commonly used facial emojis through dynamic facial and head movements while providing open-ended textual descriptions of the meanings they associated with each emoji. In the \textbf{Perception Phase}, these recorded expressions were evaluated by a separate group of perceivers, who rated how well each performance matched the associated semantic labels.

\subsection{Data Collection (Expression Phase)}
\label{subsec:dataset_expression}
\begin{figure}
\centering
  \includegraphics[width=\linewidth]{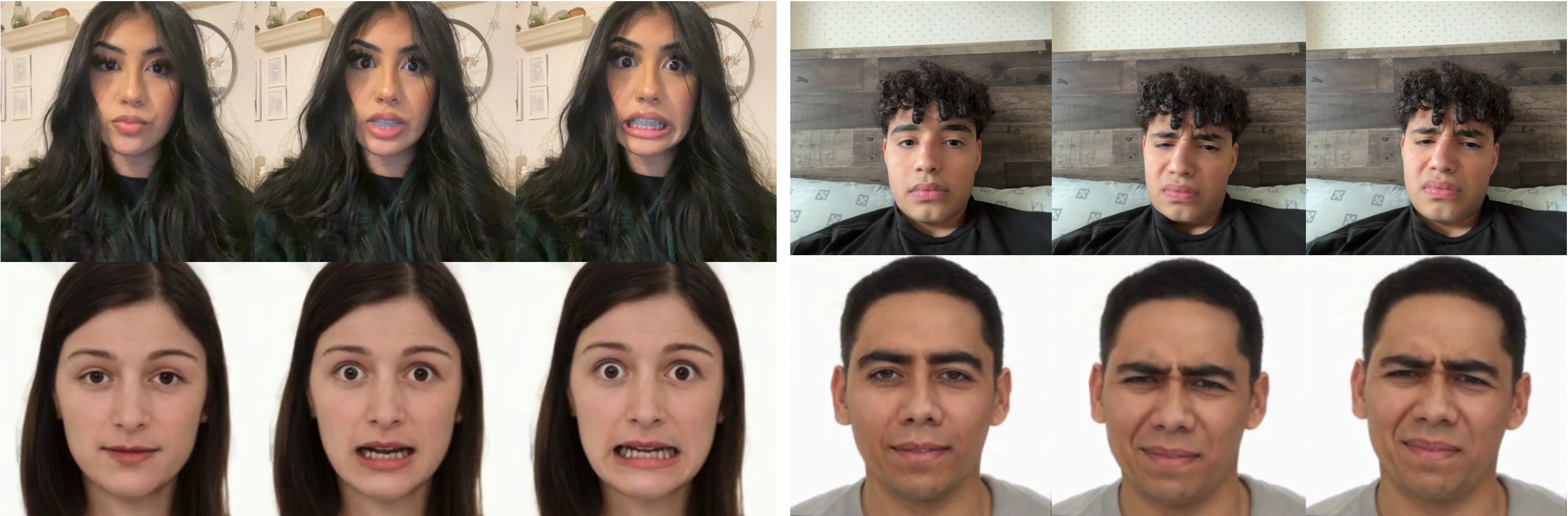}
    \caption{The first row shows the original videos recorded by participants. The second row shows the same facial motion mapped to a synthetic face using LivePortrait~\cite{guo2024liveportrait}. (The participants shown in this figure provided consent for their images to be displayed.)}
    \label{fig:livepo}
\end{figure}

\begin{figure}
  \centering
  \begin{subfigure}{0.33\columnwidth}
    \centering
    \includegraphics[
      width=\linewidth,
      keepaspectratio]{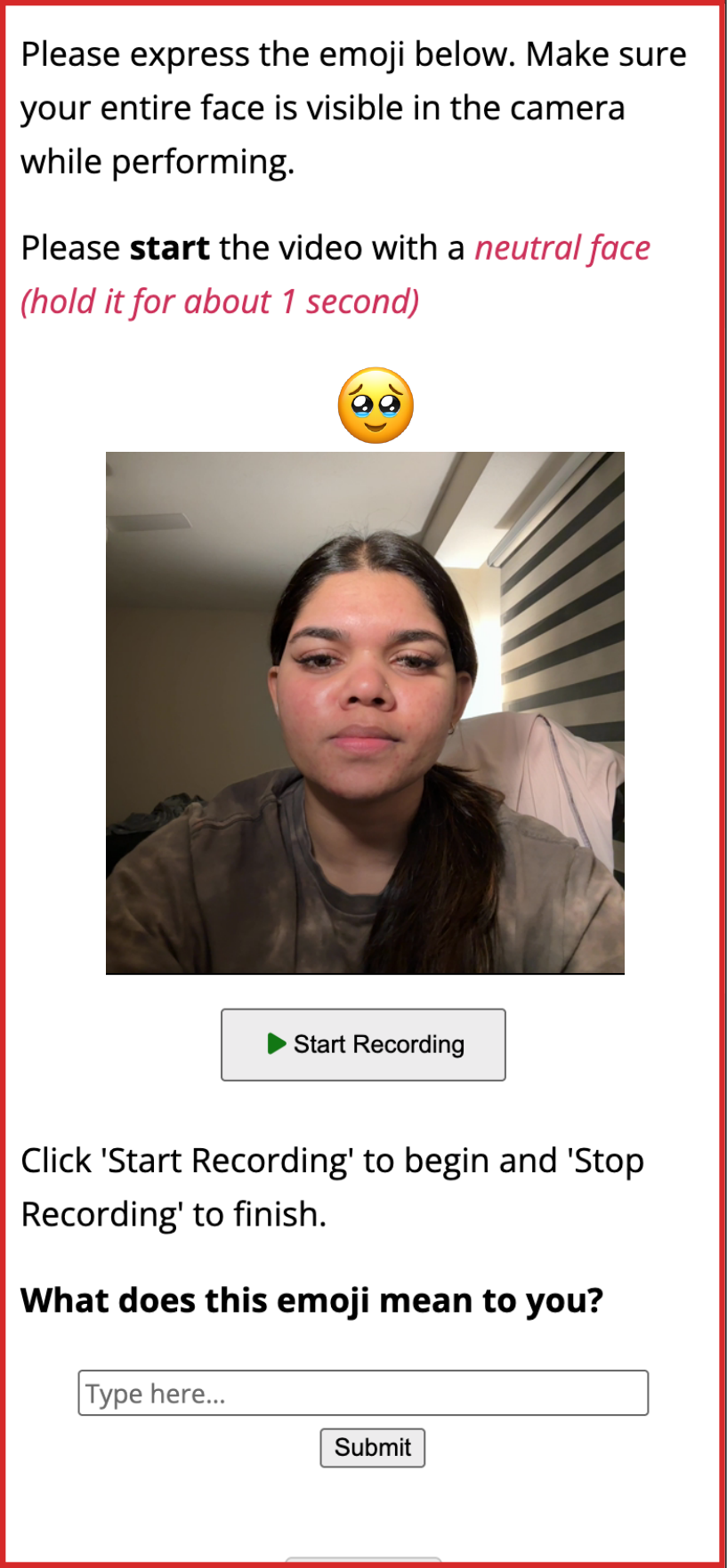}
    \caption{}
    \label{perform_screenshot}
  \end{subfigure}
  \hfill
  \begin{subfigure}{0.655\columnwidth}
    \centering
    \includegraphics[
      width=\linewidth,
      keepaspectratio]{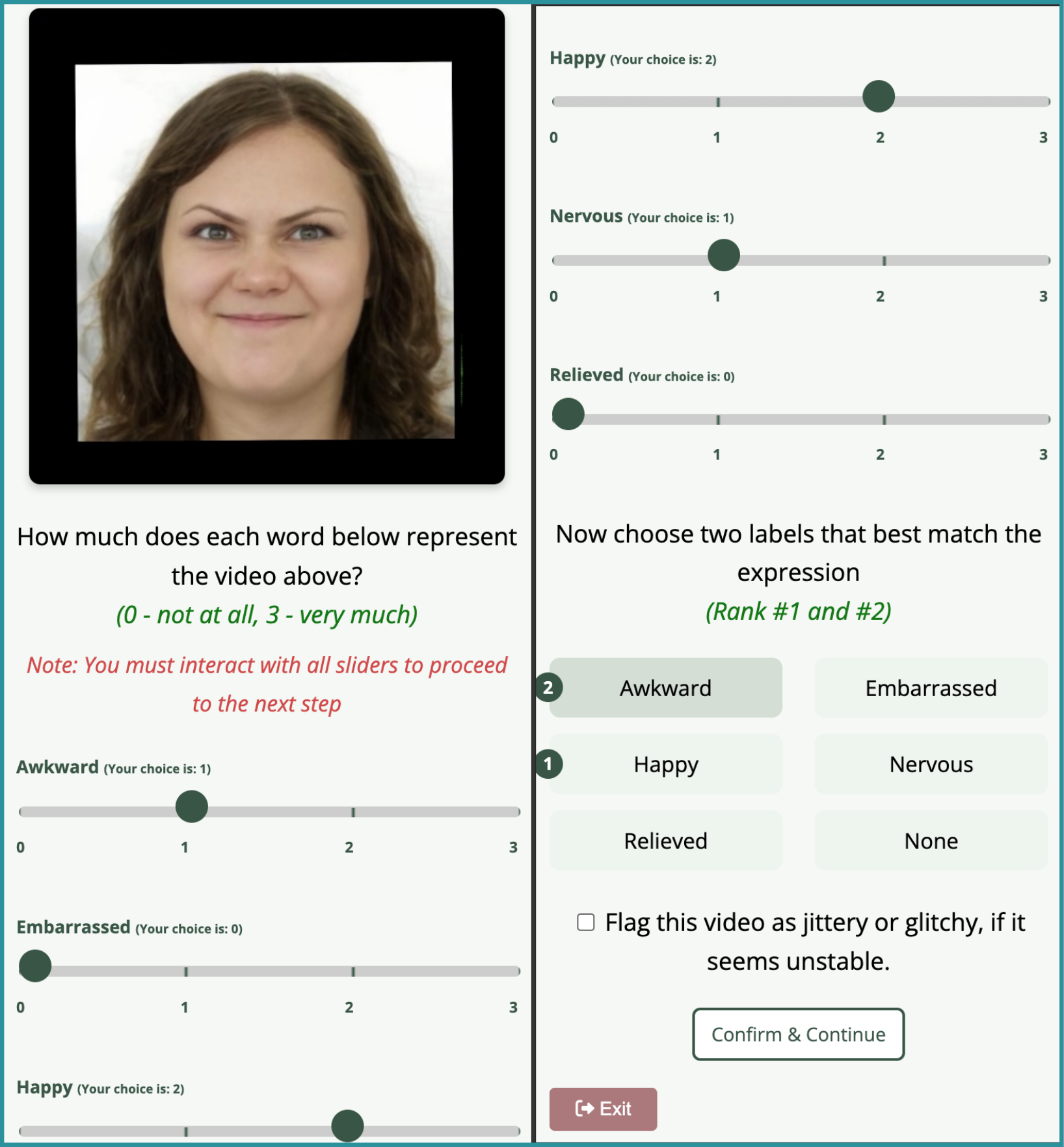}
    \caption{}
    \label{perceive_screenshot}
  \end{subfigure}
  \caption{Screenshots of (a) the data collection interface: participants were prompted with an emoji to record a facial expression video, and (b) the data validation interfaces: participants viewed anonymized videos and annotated them. (The participant shown in this figure (a) provided consent for their image to be displayed.)}
  \label{fig:two-phone-screenshots}
\end{figure}

In the expression phase, we recruited 280 participants from a university in North America, of whom 203 completed the task (45 male, 156 female, 2 unreported; mean age = 20.04) to record facial expressions corresponding to 40 commonly used facial emojis. 
Participants were instructed to express each emoji on video using natural facial and head movements, starting from a neutral expression, transitioning to the expression (Figure \ref{fig:livepo} top row). In addition to the video recordings, participants provided open-ended textual descriptions of what each emoji meant to them, allowing unconstrained semantic interpretations to be collected alongside expressive behavior. Data collection was conducted through a web-based recording interface (Figure~\ref{perform_screenshot}).

A total of \textbf{11,055 videos} were recorded. They were divided among the research team to review for adherence to the recording instructions and overall quality: videos with issues such as improper camera positioning, occlusions, incomplete expressions, or insufficient visibility of facial movement were removed. After this cleaning process, the dataset retained 2,400 facial expression videos, each capturing a complete dynamic expression sequence. Open-ended semantic descriptions were retained for all participants, including for unsuccessful video-recording attempts, resulting in a total of 11,164 valid text descriptions used in subsequent analyses, described in Section~\ref{subsec:dataset_semantic_label}.
\subsection{Expression Mapping and Anonymization}
\label{subsec:dataset_face_anonymization}
To preserve participant privacy and ensure visual consistency across the dataset, all 2,400 expression videos were anonymized. We generated a set of photorealistic base identities using the Chicago Face Database \cite{ma2015chicago}. Facial components from images in \cite{ma2015chicago} including the base face, eye region, and mouth region were composited from different source images within the dataset to disrupt identity cues. Each patched composite face was then projected into the StyleGAN2 latent space, and the nearest latent was used to generate a visually coherent synthetic identity \cite{karras2020analyzing}. EmoStyle \cite{azari2024emostyle} was used to produce neutral baseline facial appearances. This process resulted in a set of 42 synthetic individuals. 
In addition, a consistent white background across videos helped unify the dataset and reduce visual variability unrelated to facial expression.

Facial motion from the original expression videos was transferred onto the generated identities using LivePortrait \cite{guo2024liveportrait}. Two samples are shown in Figure~\ref{fig:livepo}. This process produced anonymized dynamic facial expression videos used for downstream perceptual annotation. 
To assess motion preservation, we compared MediaPipe facial-landmark displacements between each original video and its reenacted video. Across all videos, the mean cosine similarity was 0.958, suggesting strong preservation of facial-landmark motion. This does not guarantee lossless reenactment; all annotations were collected directly from the reenacted videos.
When assigning synthetic identities to performers, we aimed to approximately match gender and apparent age where possible using the age/gender estimation tool from DeepFace \cite{serengil2026boosted}. This matching was intended to maintain visual plausibility while prioritizing anonymization and privacy. 
\subsection{Semantic Label Selection}
\label{subsec:dataset_semantic_label}

We created a compact set of semantic labels related to each emoji prompt using the open-ended text descriptions provided in response to ``What does this emoji mean to you?", using a protocol based on the Consensual Qualitative Research (CQR) method \cite{hill2021essentials}. The procedure was as follows: First, (1) the description dataset was  sorted by description frequency, then (2) descriptions were merged based on similar semantic meaning. Finally, (3) the top descriptions up to a maximum of six categories were selected. All categories were single-word labels that necessarily appeared in the dataset, i.e. no new labels were permitted to be introduced. Steps (2) and (3) were performed jointly by three members of the research team who were Native or fluent in English using our discussion-until-agreement procedure~\cite{hill2021essentials}. These labels were used as candidate labels in the perception phase for each video. We measured pairwise distances between description embeddings, showing that some emojis elicited greater semantic variation than others (see Supp. Material). The original descriptions are released with the dataset. 
\subsection{Data Validation (Perception Phase)}
\label{subsec:dataset_perception}
In the perception phase, we recruited \textbf{1,242 participants} (361 male, 862 female, 13 other, 6 unreported; mean age = 20.05) from a university in North America to evaluate the recorded expression videos. The sample is concentrated in a narrow age range and is gender imbalanced. The resulting annotations therefore capture variation in this sample rather than population-wide perceptual patterns. Nevertheless, the variation observed within this relatively homogeneous sample provides a test of whether models can capture differences among annotators. Perceivers were presented with anonymized videos generated through the expression mapping process and were asked to assess how well each video matched its associated semantic labels derived from Section \ref{subsec:dataset_semantic_label}. Data collection was conducted through a web-based experimental interface (Figure~\ref{perceive_screenshot}). All participants provided informed consent in accordance with an approved research ethics protocol prior to participation.
For each video, perceivers rated how well the video matched each associated semantic label using a discrete 4-point scale ranging from 0 (``no match'') to 3 (``perfect match'').
 In addition, participants selected their top 1 and top 2 labels that best described the expression or indicated \textit{``None''} if no label was appropriate. Videos were randomly selected for each participant, and the sets of videos presented for each annotator were not necessarily the same. Participants also had the option to flag videos that appeared visually unstable or glitchy due to the mapping procedure explained in Section \ref{subsec:dataset_face_anonymization}.
Perceivers also completed demographic questions. All participation was voluntary, and participants could withdraw at any time without penalty. The public dataset is released for research use with ratings, videos, and frame-level MediaPipe \cite{lugaresi2019mediapipe} facial landmark annotations (478 keypoints per frame). A final filtering process was performed on the 2,400 videos; we excluded videos that at least two perceivers flagged as ``jittery”, resulting in a final dataset comprised of \textbf{2,111 videos}, with an average of 30 perceivers per video.
To measure inter-rater agreement among perceivers, we computed Krippendorff's alpha for the top-1 and top-2 selected labels on all the videos. Low agreement for top-1 ($\alpha=0.305$) and top-2 labels ($\alpha=0.235$) suggests notable variation in annotators' perception, as predicted by literature.

\section{Evaluation}
\label{sec:benchmark_tasks}
We evaluate \textit{Chehre} on two tasks. Task 1 evaluates Chehre using the typical dominant expression recognition task, measuring whether a model can recover the top-1 and top-2 facial expression labels most strongly supported by human annotators. This task disregards annotator differences in perception and focuses on recovering the dominant human interpretations of the anonymized expression videos. Task 2, our newly proposed task, evaluates distributional expression recognition. Instead of assuming that each video has a single deterministic ground-truth label, we assume that different people perceive them differently and compare these human rating distributions to model-generated output rating distributions. In both tasks, models are given videos from the \textit{Chehre} dataset and a set of candidate labels, including the label \textit{``None''}, allowing models to indicate that none of the candidate expressions are clearly visible in the video.
\subsection{Task 1: Dominant Recognition}
In this task, we evaluate whether VLMs can recover the dominant human perception of facial expressions videos. 
To construct our human reference for this task, we aggregated human annotations by computing the mean ratings of each candidate label for each video. We then select the two labels with the highest mean ratings as the dominant human perception for that video.

To test VLMs, we prompt each model with the video and the same set of candidate labels shown to human annotators. The model is asked to select the top two expression labels. The label \textit{``None''} is also included in the list of candidate labels, in case the model does not detect any facial expression in the video.
Note that this task is a closed-set facial expression recognition and is not intended as an open-ended task; instead, it uses the same label set presented to human annotators for each video. This design allows us to evaluate model predictions against human annotations while controlling the label space and avoiding unrestricted text generation.
\subsubsection{Metrics}
\label{subsec:1_metric}

\textbf{Top-1}: We measure whether the model's top prediction matches the label with the highest mean human score for that video.

\textbf{Recall@2}: We measure the fraction of the top two highest-rated human labels recovered by the model's two predictions, ignoring order.

\textbf{nDCG@2}: We use ``$nDCG_{@2}$'' to evaluate whether the model ranks the dominant human labels in the correct order. We define the relevance score from the top-1 and top-2 average human mean ratings of each video \cite{jarvelin2002cumulated}. We compute nDCG over the model's two predicted labels.

\textbf{False None Rate}: We report ``False None'' rate to measure cases where a model selects \textit{None} despite human evidence of a visible expression. We count a prediction as false-none when \textit{None} appears in the model's top-2 predictions, fewer than 20\% of annotators selected \textit{None}, and at least one non-\textit{None} expression has a mean human score of $>1$.
All metrics are reported as averages across all videos.\\

\subsection{Task 2: Distributional Recognition}
In this task, we evaluate whether VLMs can recover the distribution of human perceptions for facial expressions in videos.
To construct the human reference for this task, instead of reducing annotations to a single mean rating vector for each video, we retain the rating vector provided by each annotator. This allows us to capture how much annotators differed in their interpretations.

Since the human reference consists of multiple annotator rating vectors per video, a single model response is insufficient for distributional comparison; we therefore generate multiple rating vectors per video under different prompting or decoding conditions. Rather than single-label recognition, this task tests whether VLMs can produce a range of valid and plausible rating combinations for the same video, reflecting the fact that human annotators may perceive the same expression differently.
\subsubsection{Metrics}
\label{subsec:2_metric}

\textbf{Valid Output}: ``$\text{Valid}$'' measures the fraction of model generations that return a complete and valid response. A model output is valid only if it assigns an integer score from 0 to 3 to every candidate label and does not add or remove labels.

\textbf{Label Ratings MAE}: ``$\text{MAE}$'' measures the absolute difference between the mean human rating vector and the mean model rating vector.

\textbf{Rating-Distribution JSD}: ``$\text{JSD}$'' measures whether the model captures the distribution of human Likert ratings for each candidate label. For each video and candidate label, we estimate the human and model rating distributions over the four Likert scores $\{0,1,2,3\}$. We then compute the Jensen-Shannon divergence between these two distributions for each label. Then, for each video rating, JSD is computed by averaging over candidate labels, and we record the mean of rating JSD across all the videos.

\textbf{Coverage Cov}: We adopted the Coverage metric ``$\text{Cov}$'' of Naeem et al. \cite{naeem2020reliable} to measure what fraction of human rating vectors are represented by at least one model-generated rating vector. For each human vector, we define its neighborhood radius as the Euclidean distance to its $k$-th nearest human neighbor. A human response is considered covered if at least one model rating vector falls within this neighborhood. Coverage is the fraction of human response vectors covered by the model, averaged across videos. Following \cite{naeem2020reliable}'s hyperparameter-selection procedure, we set k=5, the smallest value with expected Coverage above 0.95 for our sample sizes. Higher coverage indicates that model-generated responses capture a larger proportion of the variability observed across human raters.

\textbf{Average Pairwise Distance Error (APDE):} We adopt the APDE metric~\cite{barquero2023belfusion} to measure whether the model under random sampling and persona prompting generates a similar amount of rating diversity as human annotators. For each video, we compute the mean pairwise Euclidean distance among human rating vectors and, similarly, among model-generated rating vectors. APDE is the absolute difference between the human and model mean pairwise distances, averaged across videos. Lower APDE indicates that the amount of diversity in model-generated responses is closer to that observed in human ratings.
Complete definitions and formulas for the metrics used in both tasks are provided in the Suppl. Material (Sec. B.1).
\begin{figure*}[t]
\centering
  \includegraphics[width=\textwidth]{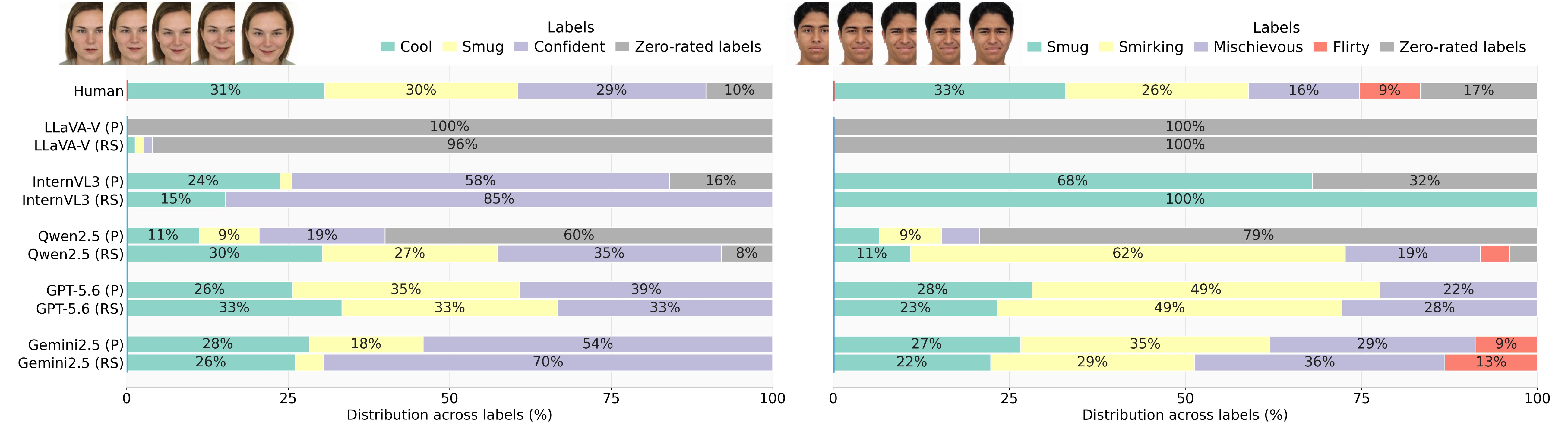}
    \caption{Distributional comparison between human ratings and model-generated outputs for two sample videos from \textit{Chehre}, with the human rating distribution (top row) and VLM-generated distributions below. Model outputs are produced using persona prompting (P) or random sampling (RS). Each stacked bar shows the percentage of ratings assigned to the candidate labels. }
    \label{fig:dist}
\end{figure*}
\section{Experiments}
\label{sec:exp}
We evaluate a number of vision-language models on our proposed tasks.


\textbf{Model selection:}
We evaluate Qwen2.5-Omni-7B \cite{xu2025qwen25omnitechnicalreport}, LLaVA-Video-7B-Qwen2 \cite{zhang2025llavavideovideoinstructiontuning}, InternVL3-8B, InternVL3-38B \cite{zhu2025internvl3}, Gemini 2.5 Flash \cite{comanici2025gemini}, GPT-5.6-Terra~\cite{openai2026gpt56}. We selected the open-source models from the OpenVLM Video Leaderboard, which reports video-understanding benchmark results using VLMEvalKit~\citep{duan2024vlmevalkit}.\footnote{\url{https://huggingface.co/spaces/opencompass/openvlm_video_leaderboard}}. We also tested Gemini 2.5 Flash (model ID: \texttt{gemini-2.5-flash}) through the Google Gemini API and GPT-5.6-Terra (model ID:  \texttt{gpt-5.6-terra} with reasoning effort set to \texttt{none}.) through the OpenAI Responses API. All API experiments were conducted in July and August 2026. The local models were evaluated on NVIDIA L40S GPUs with 48GB memory.
For all experiments, we uniformly sampled 32 frames from each video. Because GPT-5.6-Terra does not support native video input, we used the frames as temporally ordered images. Due to computational constraints, InternVL3-38B received 24 uniformly sampled frames and was evaluated only in Task 1.

\textbf{Experiment 1: Dominant recognition}
For each video, models are given the same set of candidate labels used in the perception phase task, including \textit{None}, as described in Section \ref{subsec:dataset_perception}. The model is prompted to predict the top-1 and top-2 expression labels that best describe the visible facial expression in the video. We then compare the model's predictions against the labels with the highest and second-highest mean human ratings using the metrics from Section \ref{subsec:1_metric}.
\textbf{Experiment 2: Distributional recognition}
As described in Section \ref{subsec:dataset_perception}, participants directly selected the top-1 and top-2 expression labels for each video, with \textit{None} available as an option. We use these rank selections to measure ambiguity for each video. 
For each video, we compute the entropy of the selected top-1 and top-2 label distributions, normalized by the number of candidate labels. 
We combine the top-1 and top-2 entropy scores using a weighted average, giving higher weight to top-1 selections: $A(v) = 0.75 H^{\mathrm{norm}}_1(v) + 0.25 H^{\mathrm{norm}}_2(v)$.


We then rank videos by this ambiguity score and select the top 300 videos as the high-ambiguity subset and the bottom 300 videos as the low-ambiguity subset. Since Task 2 requires multiple inference runs per video to compare VLM and human rating distributions, evaluating the full dataset is computationally expensive. We therefore report the distributional expression task on the high- and low-ambiguity subsets. 
To generate multiple model-generated rating vectors per video, we use two different settings:

First, we use \textbf{Random Sampling} implemented by stochastic decoding: each video is evaluated 25 times with the same prompt but with different random seeds ($\text{temperature}=0.7, \text{top-p}=0.9, \text{and top-k} = 20$).
\begin{table}[!b]
\centering
\small
\resizebox{\columnwidth}{!}{%
\begin{tabular}{l c c c c}
\hline
\textbf{Model Name} &  \textbf{Top-1 $\uparrow$} & \textbf{Recall@2 $\uparrow$} & \textbf{nDCG@2$\uparrow$} & \textbf{False None $\downarrow$ }  \\
\hline
LLaVA-Video-7B & 0.209 & 0.215 & 0.408 & 0.596 \\
InternVL3-8B  & 0.314 & 0.281 &  0.494 & 0.600  \\
InternVL3-38B  & 0.323 & 0.402 &  0.614 & 0.388  \\
Qwen2.5-Omni-7B & 0.332 & 0.429 &  0.664 & 0.292  \\
GPT-5.6-Terra & 0.427 & 0.590 & 0.801 & 0.182  \\
Gemini-2.5-Flash & \textbf{0.435} & \textbf{0.607} & \textbf{0.807} & \textbf{0.034}  \\
\hline
\end{tabular}%
}
\caption{Dominant expression recognition results  on the full set of 2,111 videos. Gemini-2.5-Flash achieves the best performance; overall accuracy remains low.}
\label{tab:task1_results}
\end{table}

Second, we use \textbf{Persona Prompting}: each video is evaluated 25 times with the same prompt but with a different observer perspective. This setting is inspired by the interpersonal circumplex \cite{leary2004interpersonal}, which represents interpersonal traits along two orthogonal axes: affiliation, ranging from \textit{very unfriendly to very friendly}, and dominance, ranging from \textit{very submissive to very dominant}. We use these two dimensions to define a $5 \times 5$ set of personas, resulting in 25 prompts per video.
For each persona prompt, the video, the candidate label set, and the instructions in the prompt remain fixed. Only a short annotator-perspective sentence is changed, for example:
\textit{``You are someone who is \underline{\textit{very friendly}} towards people and is \underline{\textit{submissive}}.''}
This setting evaluates whether structured observer variation can induce the model to generate more diverse scores and better match the disagreement observed in human annotators. Since persona prompting uses greedy decoding for each persona prompt, we can isolate the effect of the observer perspective changes from the effect of stochastic decoding. We also report greedy decoding without persona variation as a deterministic baseline.
Additional implementation details and the full prompts are provided in the Suppl. Material (Sec. B).

\begin{table*}
\centering
\small
\resizebox{\textwidth}{!}{%
\begin{tabular}{l l c c c c c c|c c c}
\hline
\textbf{Subset} & \textbf{Model} & \textbf{Method}
& \textbf{Valid} $\uparrow$
& \textbf{MAE} $\downarrow$
& \textbf{JSD} $\downarrow$
& \textbf{Cov.} $\uparrow$
& \textbf{APDE} $\downarrow$
&\textbf{Top-1 $\uparrow$}
&\textbf{Recall@2 $\uparrow$}
&\textbf{nDCG@2 $\uparrow$} \\
\hline


\rowcolor{yellow!8}
\cellcolor{white} & LLaVA-Video-7B & GD & 1.00 & 1.18 & 0.33 & 0.25 & -- & 0.12 & 0.44 & 0.78\\
\rowcolor{yellow!8}
\cellcolor{white} & LLaVA-Video-7B & RS & 1.00 & 1.14 & 0.32 & 0.29 & 0.95 & 0.14 & 0.46 & 0.80\\
\rowcolor{yellow!8}
\cellcolor{white} & LLaVA-Video-7B & P & 1.00 & 1.17 & 0.33 & 0.26 & 0.97 & 0.11 & 0.44 & 0.78 \\

\rowcolor{red!8}
\cellcolor{white} & InternVL3-8B & GD & 1.00 & 0.98 & 0.33 & 0.23 & -- & 0.25 & 0.51 & 0.83\\
\rowcolor{red!8}
\cellcolor{white} & InternVL3-8B & RS & 1.00 & 0.97 & 0.31 & 0.27 & 1.01 & 0.27 & 0.53 & 0.84\\
\rowcolor{red!8}
\cellcolor{white} & InternVL3-8B & P & 1.00 & 0.91 & 0.28 & 0.36 & 0.88 & 0.34 & 0.54 & 0.85 \\

\rowcolor{blue!8}
\cellcolor{white} & Qwen2.5-Omni-7B & GD & 0.94 & 0.84 & 0.34 & 0.27 & -- & 0.20 & 0.51 & 0.83\\
\rowcolor{blue!8}
\cellcolor{white} & Qwen2.5-Omni-7B & RS & 0.94 & \textbf{0.57} & \textbf{0.14} & \textbf{0.76} & \textbf{0.46} & 0.37 & 0.61 & \underline{0.89}\\
\rowcolor{blue!8}
\cellcolor{white} & Qwen2.5-Omni-7B & P & 0.94 & 0.74 & 0.20 & 0.58 & \underline{0.62} & 0.26 & 0.56 & 0.86 \\

\rowcolor{magenta!8}
\cellcolor{white} & GPT-5.6-Terra & GD & 1.00 & 0.70 & 0.33 & 0.28 & -- & 0.33 & 0.60 & 0.88\\
\rowcolor{magenta!8}
\cellcolor{white} & GPT-5.6-Terra & RS & 1.00 & 0.67 & 0.29 & 0.40 & 1.06 & \underline{0.40} & \textbf{0.64} & 0.90\\
\rowcolor{magenta!8}
\cellcolor{white} & GPT-5.6-Terra & P & 1.00 & \underline{0.63} & 0.23 & 0.56 & 0.90 & \textbf{0.44} & \textbf{0.64} & \textbf{0.91} \\

\rowcolor{green!8}
\cellcolor{white} & Gemini-2.5-Flash & GD & 1.00 & 0.88 & 0.34 & 0.25 & -- & 0.27 & 0.59 & 0.87\\
\rowcolor{green!8}
\cellcolor{white} & Gemini-2.5-Flash & RS & 1.00 & 0.78 & 0.21 & 0.56 & 0.78 & 0.39 & \underline{0.63} & \textbf{0.90}\\
\rowcolor{green!8}
\multirow{-15}{*}{\cellcolor{white}\textcolor{black}{\makecell{High\\Ambiguity\\Subset}}}
& Gemini-2.5-Flash & P & 1.00 & 0.72 & \underline{0.19} & \underline{0.63} & 0.67 & 0.38 & \underline{0.63} & \textbf{0.90} \\

\hline


\rowcolor{yellow!8}
\cellcolor{white} & LLaVA-Video-7B & GD & 1.00 & 1.32 & 0.37 & 0.11 & -- & 0.30 & 0.43 & 0.64 \\
\rowcolor{yellow!8}
\cellcolor{white} & LLaVA-Video-7B & RS & 1.00 & 1.23 & 0.32 & 0.28 & 0.77 & 0.40 & 0.61 & 0.78 \\
\rowcolor{yellow!8}
\cellcolor{white} & LLaVA-Video-7B & P & 1.00 & 1.32 & 0.37 & 0.12 & 0.92 & 0.31 & 0.43 & 0.64\\

\rowcolor{red!8}
\cellcolor{white} & InternVL3-8B & GD & 1.00 & 0.95 & 0.33 & 0.19 & -- & 0.45 & 0.63 & 0.80\\
\rowcolor{red!8}
\cellcolor{white} & InternVL3-8B & RS & 1.00 & 0.90 & 0.29 & 0.32 & 0.71 & 0.42 & 0.70 & 0.84\\
\rowcolor{red!8}
\cellcolor{white} & InternVL3-8B & P & 1.00 & 0.92 & 0.30 & 0.28 & 0.76 & 0.41 & 0.65 & 0.81 \\

\rowcolor{blue!8}
\cellcolor{white} & Qwen2.5-Omni-7B & GD & 0.95 & 0.81 & 0.32 & 0.23 & -- & 0.52 & 0.68 & 0.85 \\
\rowcolor{blue!8}
\cellcolor{white} & Qwen2.5-Omni-7B & RS & 0.95 & 0.62 & \textbf{0.15} & \underline{0.65} & \textbf{0.28} & 0.48 & 0.72 & 0.86\\
\rowcolor{blue!8}
\cellcolor{white} & Qwen2.5-Omni-7B & P & 0.95 & 0.79 & 0.23 & 0.45 & \underline{0.51} & 0.52 & 0.71 & 0.86 \\

\rowcolor{magenta!8}
\cellcolor{white} & GPT-5.6-Terra & GD & 1.00 & 0.68 & 0.30 & 0.26 & -- & \underline{0.59} & 0.75 & 0.89\\
\rowcolor{magenta!8}
\cellcolor{white} & GPT-5.6-Terra & RS & 1.00 & 0.67 & 0.27 & 0.36 & 0.89 & \underline{0.59} & 0.76 & 0.89\\
\rowcolor{magenta!8}
\cellcolor{white} & GPT-5.6-Terra & P & 1.00 & \underline{0.61} & 0.21 & 0.53 & 0.73 & 0.57 & 0.77 & \underline{0.90} \\

\rowcolor{green!8}
\cellcolor{white} & Gemini-2.5-Flash & GD & 1.00 & 0.69 & 0.29 & 0.28 & -- & \textbf{0.60} & 0.75 & 0.88\\
\rowcolor{green!8}
\cellcolor{white} & Gemini-2.5-Flash & RS & 1.00 & 0.64 & 0.18 & 0.59 & 0.65 & 0.55 & \underline{0.78} & \underline{0.90} \\
\rowcolor{green!8}
\multirow{-15}{*}{\cellcolor{white}\textcolor{black}{\makecell{Low\\Ambiguity\\Subset}}}
& Gemini-2.5-Flash & P & 1.00 & \textbf{0.58} & \underline{0.16} & \textbf{0.67} & 0.56 & 0.56 & \textbf{0.80} & \textbf{0.91} \\

\hline
\end{tabular}%
}

\caption{
Distributional expression recognition task results. Comparing greedy decoding (GD), persona prompting (P, 25 generations), and random sampling (RS, 25 generations). For each subset, best in \textbf{bold}, second best \underline{underlined}.
On high ambiguity, Qwen2.5-Omni (RS) is strongest on the distributional metrics (MAE, JSD, Cov., APDE), while GPT-5.6-Terra (P) leads dominant recognition (Top-1, Recall@2).
On low ambiguity, Gemini-2.5-Flash (P) performs best on MAE and Cov., while Qwen2.5-Omni (RS) performs best on JSD and APDE.
All methods have $\text{APDE}>0$,
suggesting a remaining mismatch between model and human response diversity. Averaging over generations (RS/P vs.\ GD) improves dominant recognition results for Qwen2.5-Omni, InternVL3, Gemini-2.5-Flash, and GPT-5.6-Terra; it does not help LLaVA-Video, which overuses the zero rating.
}
\label{tab:task2_results_most}
\end{table*}

\section{Results}
\textbf{Can models recover the dominant human interpretation of each video?} 

Table~\ref{tab:task1_results} demonstrates that the evaluated VLMs do not reliably capture the dominant human perception of the facial expression in \textit{Chehre} videos. Gemini 2.5 Flash achieves the best performance in all metrics with 43.5\% top-1 accuracy,  GPT-5.6-Terra is the second-best performing model. Notably, Gemini generates fewer false-None predictions (3.4\%) than GPT-5.6-Terra (18.2\%) and the open-source models (29.2--60.0\%).

We observe a high False None Rate for open-source models, which shows that these VLMs often choose ``\textit{None}'', even when human annotators see an expression in the video. Inspecting the outputs, models sometimes justify these ``None'' predictions with reasoning such as: \textit{``The facial expression appears neutral with no clear indication of any specific emotion.''} Table \ref{tab:task2_results_most} reports lower scores on high-ambiguity videos.
Overall, the results suggest that the selected VLMs cannot fully recover the top two labels from a fixed subset of labels in \textit{Chehre}, and the task proved challenging even for the larger-scale model tested.

\textbf{Can the selected models capture the human rating distribution?}
Table~\ref{tab:task2_results_most} reports the distributional facial expression recognition task results on the 300-video high-ambiguity and low-ambiguity subsets. Overall, the results suggest that model-generated ratings in our selected models do not fully capture the human reference variance. Figure~\ref{fig:dist} provides a visual comparison of the human and model-generated rating distributions for two example videos.
On the high-ambiguity subset, Qwen2.5-Omni-7B with random sampling covers the human distribution better than all other models. Gemini-2.5-Flash with persona prompting achieves the second-highest Coverage (0.63), and GPT-5.6-Terra with persona prompting obtains the second-lowest MAE (0.63). LLaVA-Video performs poorly on both subsets, with high MAE and JSD, low Coverage, and high APDE, suggesting that its generated ratings are poorly aligned with the human distributions and cover only a limited portion of the variation observed in human annotators. 
On the low-ambiguity subset, Gemini-2.5-Flash with persona prompting achieves the lowest MAE (0.58) and highest Coverage (0.67), and Qwen2.5-Omni-7B with random sampling achieves the lowest JSD (0.15) and APDE (0.28). These results suggest that the low-ambiguity subset does not eliminate the gap between model-generated and human rating distributions.

\textbf{Does persona prompting introduce variation?}
Compared with greedy decoding, persona prompting increases Coverage for every model on both subsets, suggesting that giving the models different personas generates outputs that cover a larger portion of the human rating distribution. However, whether persona prompting performs better than random sampling is model-dependent. In the top-performing models, persona prompting improves all four distributional metrics over random sampling for GPT-5.6-Terra and Gemini-2.5-Flash on both subsets, whereas Qwen2.5-Omni-7B performs better with random sampling. These results suggest that persona prompting can act as a controllable way to shift model perception and better capture human response variation, but its effectiveness depends on the model. Figure~\ref{fig:persona} shows examples of outputs generated under different persona prompts for Qwen2.5-Omni-7B.

\textbf{Does capturing variation help with dominant-label prediction?}
Table~\ref{tab:task2_results_most} (right side) reports the Task 1 metrics on the same high-  and low-ambiguity subsets. Interestingly, on the high-ambiguity subset, averaging over multiple runs through random sampling or persona settings achieves higher results in the Dominant Expression Recognition task (top1 and top2) compared to vanilla greedy decoding, suggesting that modeling variation is a promising approach even for the typical task of most common label recognition. This pattern is particularly clear for Qwen2.5-Omni-7B, InternVL3-8B, GPT-5.6-Terra, and Gemini-2.5-Flash.
On the low-ambiguity subset, the effect is weaker; multiple runs generally improve Recall@2 and nDCG@2, but do not consistently improve Top-1 accuracy. These results suggest that capturing variation may be more useful when human interpretations are themselves more variable.

\section{Future Work}
Future work can include evaluations of a broader range of VLMs, and studies of how perception varies based on annotator demographics. In addition, we plan to further explore persona prompting as a controllable way to induce variation in VLM predictions~\cite{calder2011personality}, including alternative persona constructions. Another direction is to examine how the choice of candidate labels for each video affects human/model perception and to compare closed-set labels with more open-ended descriptions of facial expressions. Finally, \textit{Chehre} can support future studies on the effect of the apparent gender, ethnicity, or age of the synthetic face on the perceived facial expression.

{
    \small
    \bibliographystyle{ieeenat_fullname}
    \bibliography{main}

@String(ICCV= {Int. Conf. Comput. Vis.})

@String(ICCV  = {ICCV})

@article{lugaresi2019mediapipe,
  title={Mediapipe: A framework for building perception pipelines},
  author={Lugaresi, Camillo and Tang, Jiuqiang and Nash, Hadon and McClanahan, Chris and Uboweja, Esha and Hays, Michael and Zhang, Fan and Chang, Chuo-Ling and Yong, Ming Guang and Lee, Juhyun and others},
  journal={arXiv preprint arXiv:1906.08172},
  year={2019}
}

@article{guo2024liveportrait,
  title={Liveportrait: Efficient portrait animation with stitching and retargeting control},
  author={Guo, Jianzhu and Zhang, Dingyun and Liu, Xiaoqiang and Zhong, Zhizhou and Zhang, Yuan and Wan, Pengfei and Zhang, Di},
  journal={arXiv preprint arXiv:2407.03168},
  year={2024}
}

@inproceedings{azari2024emostyle,
  title={Emostyle: One-shot facial expression editing using continuous emotion parameters},
  author={Azari, Bita and Lim, Angelica},
  booktitle={Proceedings of the IEEE/CVF Winter Conference on Applications of Computer Vision},
  pages={6385--6394},
  year={2024}
}

@article{ma2015chicago,
  title={The Chicago face database: A free stimulus set of faces and norming data},
  author={Ma, Debbie S and Correll, Joshua and Wittenbrink, Bernd},
  journal={Behavior research methods},
  volume={47},
  number={4},
  pages={1122--1135},
  year={2015},
  publisher={Springer}
}

@article{jarvelin2002cumulated,
  title={Cumulated gain-based evaluation of IR techniques},
  author={J{\"a}rvelin, Kalervo and Kek{\"a}l{\"a}inen, Jaana},
  journal={ACM Transactions on Information Systems (TOIS)},
  volume={20},
  number={4},
  pages={422--446},
  year={2002},
  publisher={ACM New York, NY, USA}
}

@inproceedings{duan2024vlmevalkit,
  title={Vlmevalkit: An open-source toolkit for evaluating large multi-modality models},
  author={Duan, Haodong and Yang, Junming and Qiao, Yuxuan and Fang, Xinyu and Chen, Lin and Liu, Yuan and Dong, Xiaoyi and Zang, Yuhang and Zhang, Pan and Wang, Jiaqi and others},
  booktitle={Proceedings of the 32nd ACM International Conference on Multimedia},
  pages={11198--11201},
  year={2024}
}

@article{zhu2025internvl3,
  title={Internvl3: Exploring advanced training and test-time recipes for open-source multimodal models},
  author={Zhu, Jinguo and Wang, Weiyun and Chen, Zhe and Liu, Zhaoyang and Ye, Shenglong and Gu, Lixin and Tian, Hao and Duan, Yuchen and Su, Weijie and Shao, Jie and others},
  journal={arXiv preprint arXiv:2504.10479},
  year={2025}
}

@misc{zhang2025llavavideovideoinstructiontuning,
      title={LLaVA-Video: Video Instruction Tuning With Synthetic Data}, 
      author={Yuanhan Zhang and Jinming Wu and Wei Li and Bo Li and Zejun Ma and Ziwei Liu and Chunyuan Li},
      year={2025},
      eprint={2410.02713},
      archivePrefix={arXiv},
      primaryClass={cs.CV},
      url={https://arxiv.org/abs/2410.02713}, 
}

@book{leary2004interpersonal,
  title={Interpersonal diagnosis of personality: A functional theory and methodology for personality evaluation},
  author={Leary, Timothy},
  year={2004},
  publisher={Wipf and Stock Publishers}
}

@article{mohanty2025top,
  title={Top-down influences on the perception of emotional stimuli},
  author={Mohanty, Aprajita and Freeman, Jonathan and Jin, Jingwen},
  journal={Nature Reviews Psychology},
  volume={4},
  number={6},
  pages={388--403},
  year={2025},
  publisher={Nature Publishing Group US New York}
}

@article{ambadar2005deciphering,
  title={Deciphering the enigmatic face: The importance of facial dynamics in interpreting subtle facial expressions},
  author={Ambadar, Zara and Schooler, Jonathan W and Cohn, Jeffrey F},
  journal={Psychological science},
  volume={16},
  number={5},
  pages={403--410},
  year={2005},
  publisher={SAGE Publications Sage CA: Los Angeles, CA}
}

@article{cowen2020face,
  title={What the face displays: Mapping 28 emotions conveyed by naturalistic expression.},
  author={Cowen, Alan S and Keltner, Dacher},
  journal={American Psychologist},
  volume={75},
  number={3},
  pages={349},
  year={2020},
  publisher={American Psychological Association}
}

@article{kollias2018aff,
  title={Aff-wild2: Extending the aff-wild database for affect recognition},
  author={Kollias, Dimitrios and Zafeiriou, Stefanos},
  journal={arXiv preprint arXiv:1811.07770},
  year={2018}
}

@inproceedings{ren2024veatic,
  title={Veatic: Video-based emotion and affect tracking in context dataset},
  author={Ren, Zhihang and Ortega, Jefferson and Wang, Yifan and Chen, Zhimin and Guo, Yunhui and Yu, Stella X and Whitney, David},
  booktitle={Proceedings of the IEEE/CVF winter conference on applications of computer vision},
  pages={4467--4477},
  year={2024}
}

@inproceedings{lee2019context,
  title={Context-aware emotion recognition networks},
  author={Lee, Jiyoung and Kim, Seungryong and Kim, Sunok and Park, Jungin and Sohn, Kwanghoon},
  booktitle={Proceedings of the IEEE/CVF international conference on computer vision},
  pages={10143--10152},
  year={2019}
}

@inproceedings{poria2019meld,
  title={Meld: A multimodal multi-party dataset for emotion recognition in conversations},
  author={Poria, Soujanya and Hazarika, Devamanyu and Majumder, Navonil and Naik, Gautam and Cambria, Erik and Mihalcea, Rada},
  booktitle={Proceedings of the 57th annual meeting of the association for computational linguistics},
  pages={527--536},
  year={2019}
}

@article{barrett2019emotional,
  title={Emotional expressions reconsidered: Challenges to inferring emotion from human facial movements},
  author={Barrett, Lisa Feldman and Adolphs, Ralph and Marsella, Stacy and Martinez, Aleix M and Pollak, Seth D},
  journal={Psychological science in the public interest},
  volume={20},
  number={1},
  pages={1--68},
  year={2019},
  publisher={Sage Publications Sage CA: Los Angeles, CA}
}

@article{mollahosseini2017affectnet,
  title={Affectnet: A database for facial expression, valence, and arousal computing in the wild},
  author={Mollahosseini, Ali and Hasani, Behzad and Mahoor, Mohammad H},
  journal={IEEE transactions on affective computing},
  volume={10},
  number={1},
  pages={18--31},
  year={2017},
  publisher={IEEE}
}

@inproceedings{li2017reliable,
  title={Reliable crowdsourcing and deep locality-preserving learning for expression recognition in the wild},
  author={Li, Shan and Deng, Weihong and Du, JunPing},
  booktitle={Proceedings of the IEEE conference on computer vision and pattern recognition},
  pages={2852--2861},
  year={2017}
}

@article{lian2024gpt,
  title={Gpt-4v with emotion: A zero-shot benchmark for generalized emotion recognition},
  author={Lian, Zheng and Sun, Licai and Sun, Haiyang and Chen, Kang and Wen, Zhuofan and Gu, Hao and Liu, Bin and Tao, Jianhua},
  journal={Information Fusion},
  volume={108},
  pages={102367},
  year={2024},
  publisher={Elsevier}
}

@article{cheng2024emotion,
  title={Emotion-{LL}a{MA}: Multimodal emotion recognition and reasoning with instruction tuning},
  author={Cheng, Zebang and Cheng, Zhi-Qi and He, Jun-Yan and Sun, Jingdong and Wang, Kai and Lin, Yuxiang and Lian, Zheng and Peng, Xiaojiang and Hauptmann, Alexander G},
  journal={Advances in Neural Information Processing Systems},
  volume={37},
  pages={110805--110853},
  year={2024}
}

@inproceedings{sabour2024emobench,
  title={Emobench: Evaluating the emotional intelligence of large language models},
  author={Sabour, Sahand and Liu, Siyang and Zhang, Zheyuan and Liu, June and Zhou, Jinfeng and Sunaryo, Alvionna and Lee, Tatia and Mihalcea, Rada and Huang, Minlie},
  booktitle={Proceedings of the 62nd Annual Meeting of the Association for Computational Linguistics (Volume 1: Long Papers)},
  pages={5986--6004},
  year={2024}
}

@inproceedings{chakraborty-etal-2025-vibe,
    title = "{VIBE}: Can a {VLM} Read the Room?",
    author = "Chakraborty, Tania  and
      Caplan, Eylon  and
      Goldwasser, Dan",
    editor = "Christodoulopoulos, Christos  and
      Chakraborty, Tanmoy  and
      Rose, Carolyn  and
      Peng, Violet",
    booktitle = "Findings of the Association for Computational Linguistics: EMNLP 2025",
    month = nov,
    year = "2025",
    address = "Suzhou, China",
    publisher = "Association for Computational Linguistics",
    url = "https://aclanthology.org/2025.findings-emnlp.1252/",
    doi = "10.18653/v1/2025.findings-emnlp.1252",
    pages = "22992--23008",
    ISBN = "979-8-89176-335-7",
}

@article{zhang2025mme,
  title={Mme-emotion: A holistic evaluation benchmark for emotional intelligence in multimodal large language models},
  author={Zhang, Fan and Cheng, Zebang and Deng, Chong and Li, Haoxuan and Lian, Zheng and Chen, Qian and Liu, Huadai and Wang, Wen and Zhang, Yi-Fan and Zhang, Renrui and others},
  journal={arXiv preprint arXiv:2508.09210},
  year={2025}
}

@inproceedings{frohling2025personas,
  title={Personas with attitudes: Controlling llms for diverse data annotation},
  author={Fr{\"o}hling, Leon and Demartini, Gianluca and Assenmacher, Dennis},
  booktitle={Proceedings of the The 9th Workshop on Online Abuse and Harms (WOAH)},
  pages={468--481},
  year={2025}
}

@article{lutz2025prompt,
  title={The prompt makes the person (a): A systematic evaluation of sociodemographic persona prompting for large language models},
  author={Lutz, Marlene and Sen, Indira and Ahnert, Georg and Rogers, Elisa and Strohmaier, Markus},
  journal={arXiv preprint arXiv:2507.16076},
  year={2025}
}

@article{uma2021learning,
  title={Learning from disagreement: A survey},
  author={Uma, Alexandra N and Fornaciari, Tommaso and Hovy, Dirk and Paun, Silviu and Plank, Barbara and Poesio, Massimo},
  journal={Journal of Artificial Intelligence Research},
  volume={72},
  pages={1385--1470},
  year={2021}
}

@inproceedings{plank2022problem,
  title={The “problem” of human label variation: On ground truth in data, modeling and evaluation},
  author={Plank, Barbara},
  booktitle={Proceedings of the 2022 conference on empirical methods in natural language processing},
  pages={10671--10682},
  year={2022}
}

@misc{davani2021dealingdisagreementslookingmajority,
      title={Dealing with Disagreements: Looking Beyond the Majority Vote in Subjective Annotations}, 
      author={Aida Mostafazadeh Davani and Mark Díaz and Vinodkumar Prabhakaran},
      year={2021},
      eprint={2110.05719},
      archivePrefix={arXiv},
      primaryClass={cs.CL},
      url={https://arxiv.org/abs/2110.05719}, 
}

@inproceedings{ortmann2024emojiherovr,
  title={EmojiHeroVR: a study on facial expression recognition under partial occlusion from head-mounted displays},
  author={Ortmann, Thorben and Wang, Qi and Putzar, Larissa},
  booktitle={2024 12th International Conference on Affective Computing and Intelligent Interaction (ACII)},
  pages={80--88},
  year={2024},
  organization={IEEE}
}

@inproceedings{karras2020analyzing,
  title={Analyzing and improving the image quality of stylegan},
  author={Karras, Tero and Laine, Samuli and Aittala, Miika and Hellsten, Janne and Lehtinen, Jaakko and Aila, Timo},
  booktitle={Proceedings of the IEEE/CVF conference on computer vision and pattern recognition},
  pages={8110--8119},
  year={2020}
}

@article{serengil2026boosted,
  title     = {Boosted LightFace: A Hybrid DNN and GBM Model for Boosted Facial Recognition},
  author    = {Serengil, Sefik Ilkin and Ozpinar, Alper},
  journal   = {Gazi University Journal of Science},
  volume    = {39},
  number    = {1},
  pages     = {452--466},
  year      = {2026},
  doi       = {10.35378/gujs.1794891},
  url       = {https://dergipark.org.tr/en/pub/gujs/article/1794891},
  publisher = {Gazi University}
}

@inproceedings{etesam2024contextual,
  title={Contextual emotion recognition using large vision language models},
  author={Etesam, Yasaman and Yal{\c{c}}{\i}n, {\"O}zge Nilay and Zhang, Chuxuan and Lim, Angelica},
  booktitle={2024 IEEE/RSJ International Conference on Intelligent Robots and Systems (IROS)},
  pages={4769--4776},
  year={2024},
  organization={IEEE}
}

@article{fang2021cultural,
  title={Cultural differences in perceiving transitions in emotional facial expressions: Easterners show greater contrast effects than westerners},
  author={Fang, Xia and van Kleef, Gerben A and Kawakami, Kerry and Sauter, Disa A},
  journal={Journal of Experimental Social Psychology},
  volume={95},
  pages={104143},
  year={2021},
  publisher={Elsevier}
}

@article{jack2012facial,
  title={Facial expressions of emotion are not culturally universal},
  author={Jack, Rachael E and Garrod, Oliver GB and Yu, Hui and Caldara, Roberto and Schyns, Philippe G},
  journal={Proceedings of the National Academy of Sciences},
  volume={109},
  number={19},
  pages={7241--7244},
  year={2012},
  publisher={National Academy of Sciences}
}

@article{erle2022emojis,
  title={Emojis as social information in digital communication.},
  author={Erle, Thorsten M and Schmid, Karoline and Goslar, Simon H and Martin, Jared D},
  journal={Emotion},
  volume={22},
  number={7},
  pages={1529},
  year={2022},
  publisher={American Psychological Association}
}

@book{hill2021essentials,
  title={Essentials of consensual qualitative research.},
  author={Hill, Clara E and Knox, Sarah},
  year={2021},
  publisher={American Psychological Association}
}

@article{calder2011personality,
  title={Personality influences the neural responses to viewing facial expressions of emotion},
  author={Calder, Andrew J and Ewbank, Michael and Passamonti, Luca},
  journal={Philosophical Transactions of the Royal Society B: Biological Sciences},
  volume={366},
  number={1571},
  pages={1684--1701},
  year={2011},
  publisher={The Royal Society}
}

@article{binetti2022genetic,
  title={Genetic algorithms reveal profound individual differences in emotion recognition},
  author={Binetti, Nicola and Roubtsova, Nadejda and Carlisi, Christina and Cosker, Darren and Viding, Essi and Mareschal, Isabelle},
  journal={Proceedings of the National Academy of Sciences},
  volume={119},
  number={45},
  pages={e2201380119},
  year={2022},
  publisher={National Academy of Sciences}
}

@article{geng2016label,
  title={Label distribution learning},
  author={Geng, Xin},
  journal={IEEE Transactions on Knowledge and Data Engineering},
  volume={28},
  number={7},
  pages={1734--1748},
  year={2016},
  publisher={IEEE}
}

@inproceedings{jiang2020dfew,
  title={Dfew: A large-scale database for recognizing dynamic facial expressions in the wild},
  author={Jiang, Xingxun and Zong, Yuan and Zheng, Wenming and Tang, Chuangao and Xia, Wanchuang and Lu, Cheng and Liu, Jiateng},
  booktitle={Proceedings of the 28th ACM international conference on multimedia},
  pages={2881--2889},
  year={2020}
}

@inproceedings{lyons1998japanese,
  title={The Japanese female facial expression (JAFFE) database},
  author={Lyons, Michael J and Akamatsu, Shigeru and Kamachi, Miyuki and Gyoba, Jiro and Budynek, Julien},
  booktitle={Proceedings of third international conference on automatic face and gesture recognition},
  pages={14--16},
  year={1998}
}

@article{comanici2025gemini,
  title={Gemini 2.5: Pushing the frontier with advanced reasoning, multimodality, long context, and next generation agentic capabilities},
  author={Comanici, Gheorghe and Bieber, Eric and Schaekermann, Mike and Pasupat, Ice and Sachdeva, Noveen and Dhillon, Inderjit and Blistein, Marcel and Ram, Ori and Zhang, Dan and Rosen, Evan and others},
  journal={arXiv preprint arXiv:2507.06261},
  year={2025}
}

@misc{xu2025qwen25omnitechnicalreport,
      title={Qwen2.5-Omni Technical Report}, 
      author={Jin Xu and Zhifang Guo and Jinzheng He and Hangrui Hu and Ting He and Shuai Bai and Keqin Chen and Jialin Wang and Yang Fan and Kai Dang and Bin Zhang and Xiong Wang and Yunfei Chu and Junyang Lin},
      year={2025},
      eprint={2503.20215},
      archivePrefix={arXiv},
      primaryClass={cs.CL},
      url={https://arxiv.org/abs/2503.20215}, 
}

@inproceedings{naeem2020reliable,
  title={Reliable fidelity and diversity metrics for generative models},
  author={Naeem, Muhammad Ferjad and Oh, Seong Joon and Uh, Youngjung and Choi, Yunjey and Yoo, Jaejun},
  booktitle={International conference on machine learning},
  pages={7176--7185},
  year={2020},
  organization={PMLR}
}

@inproceedings{barquero2023belfusion,
  title={Belfusion: Latent diffusion for behavior-driven human motion prediction},
  author={Barquero, German and Escalera, Sergio and Palmero, Cristina},
  booktitle={2023 IEEE/CVF International Conference on Computer Vision (ICCV)},
  pages={2317--2327},
  year={2023},
  organization={IEEE}
}

@misc{openai2026gpt56,
  author       = {{OpenAI}},
  title        = {{GPT-5.6 System Card}},
  year         = {2026},
  month        = jul,
  howpublished = {\url{https://deploymentsafety.openai.com/gpt-5-6/gpt-5-6.pdf}},
  note         = {Accessed: August 20, 2026}
}

@article{chung2023unimax,
  title={Unimax: Fairer and more effective language sampling for large-scale multilingual pretraining},
  author={Chung, Hyung Won and Constant, Noah and Garcia, Xavier and Roberts, Adam and Tay, Yi and Narang, Sharan and Firat, Orhan},
  journal={arXiv preprint arXiv:2304.09151},
  year={2023}
}
}

\clearpage
\appendix

\section{Additional Dataset Details}
\label{appendix:dataset_details}

\newcommand{\emojiimg}[1]{\raisebox{-0.2ex}{\includegraphics[height=1.0em]{images/emojis/#1.png}}}

\subsection{Semantic Diversity of Text Descriptions}
\label{appendix:description_diversity}

We measured the semantic similarity among the open-ended descriptions. We encoded each text description using UMT5-XXL \cite{chung2023unimax}, mean-pooled the token embeddings, and normalized the resulting representation. For emoji $e$ with $n_e$ descriptions, we calculate the mean pairwise cosine similarity as follows:

\begin{equation}
    \mathrm{Sim}(e) =
    \frac{2}{n_e(n_e-1)}
    \sum_{i<i'}
    z_{e,i}^{\top}z_{e,i'},
\end{equation}

Lower similarity suggests that the descriptions are more semantically diverse, whereas higher similarity suggests more consistent interpretation
Table~\ref{tab:emoji_description_similarity} shows the five emojis that their input text descriptions have the lowest and highest average cosine similarity.

\subsection{Expression Phase Recording Protocol}
\label{appendix:exp_protocol}
 During the expression phase, participants were instructed to complete the recordings in a quiet environment, position their recording device (mobile phone or laptop) at eye level, look directly at the camera, and ensure that their entire face was visible within the frame. Guidance was provided on appropriate camera distance, and participants were instructed to keep their hands out of view during recording. Prior to participation, all individuals reviewed and gave informed consent in accordance with an approved research ethics protocol, which detailed the purpose of the study, video recording procedures, data handling procedures, and options for data sharing and withdrawal. As part of the data collection procedure, participants completed a demographic questionnaire that included age group, gender identity, country of birth, ancestry/ethnicity background, and marital status. Individuals received course credit for their participation.

\subsection{Synthetic Face Generation}
\label{appendix:pp_detail}
\begin{figure}
    \centering
\includegraphics[width=0.8\linewidth]{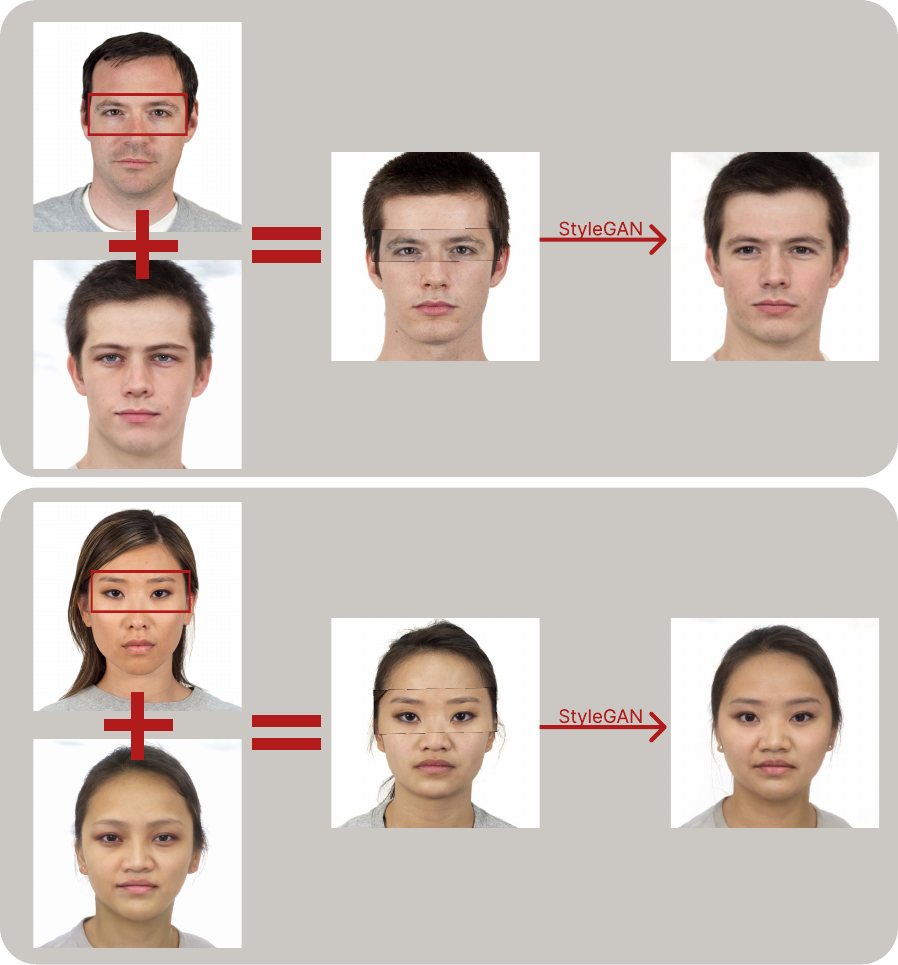}
    \caption{Two samples of generated faces. For each one two faces from the Chicago Face Database are used \cite{ma2015chicago}: the eyes of one face are patched onto the remaining regions of another face. Then each patched composite face is projected into the StyleGAN2~\cite{karras2020analyzing} latent space, and the nearest latent is used to generate a synthetic face.}
    \label{fig:synthetic_faces}
\end{figure}

Figure~\ref{fig:synthetic_faces} shows examples of the generated synthetic identities used for our study. The synthetic faces were constructed by combining facial components from Chicago Face Database images and refining the results with StyleGAN2~\cite{karras2020analyzing}. We then applied EmoStyle~\cite{azari2024emostyle} to adjust the perceived facial expression toward neutral valence and arousal.

\begin{table}[t]
\centering
\small
\setlength{\tabcolsep}{8pt}
\begin{tabular}{cc|cc}
\hline
\multicolumn{2}{c|}{\textbf{Most diverse}} &
\multicolumn{2}{c}{\textbf{Least diverse}} \\
\textbf{Emoji} & \textbf{Mean similarity} &
\textbf{Emoji} & \textbf{Mean similarity} \\
\hline
\emojiimg{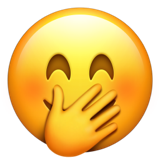} & 0.472 & \emojiimg{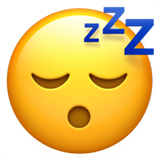} & 0.634 \\
\emojiimg{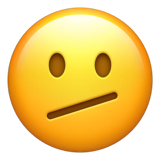} & 0.475 & \emojiimg{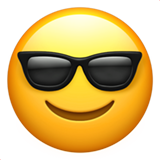} & 0.603 \\
\emojiimg{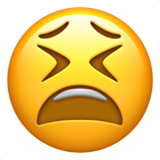} & 0.478 & \emojiimg{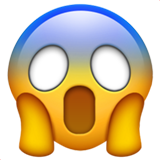} & 0.585 \\
\emojiimg{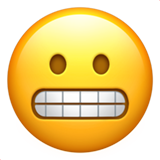} & 0.481 & \emojiimg{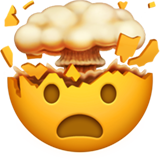} & 0.584 \\
\emojiimg{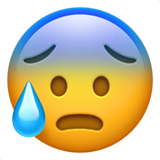} & 0.481 & \emojiimg{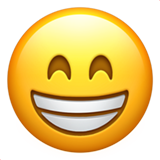} & 0.577 \\
\hline
\end{tabular}
\caption{Emojis with the lowest and highest mean pairwise cosine similarities among their open-ended descriptions. Lower similarity indicates greater semantic diversity.}
\label{tab:emoji_description_similarity}
\end{table}

\subsection{Expression/Perception Phase Questionnaire}
\label{appendix:questionnare}
Participants were recruited through a university participant pool and received course credit.
In both steps of data collection and annotation, participants after reviewing and accepting the consent form clauses, were directed to complete a questionnaire and then proceeded to the study (Expression or Perception). The questions are shown in Table~\ref{tab:demographic_questionnaire}.
\begin{figure*}
    \centering
    \includegraphics[width=\linewidth]{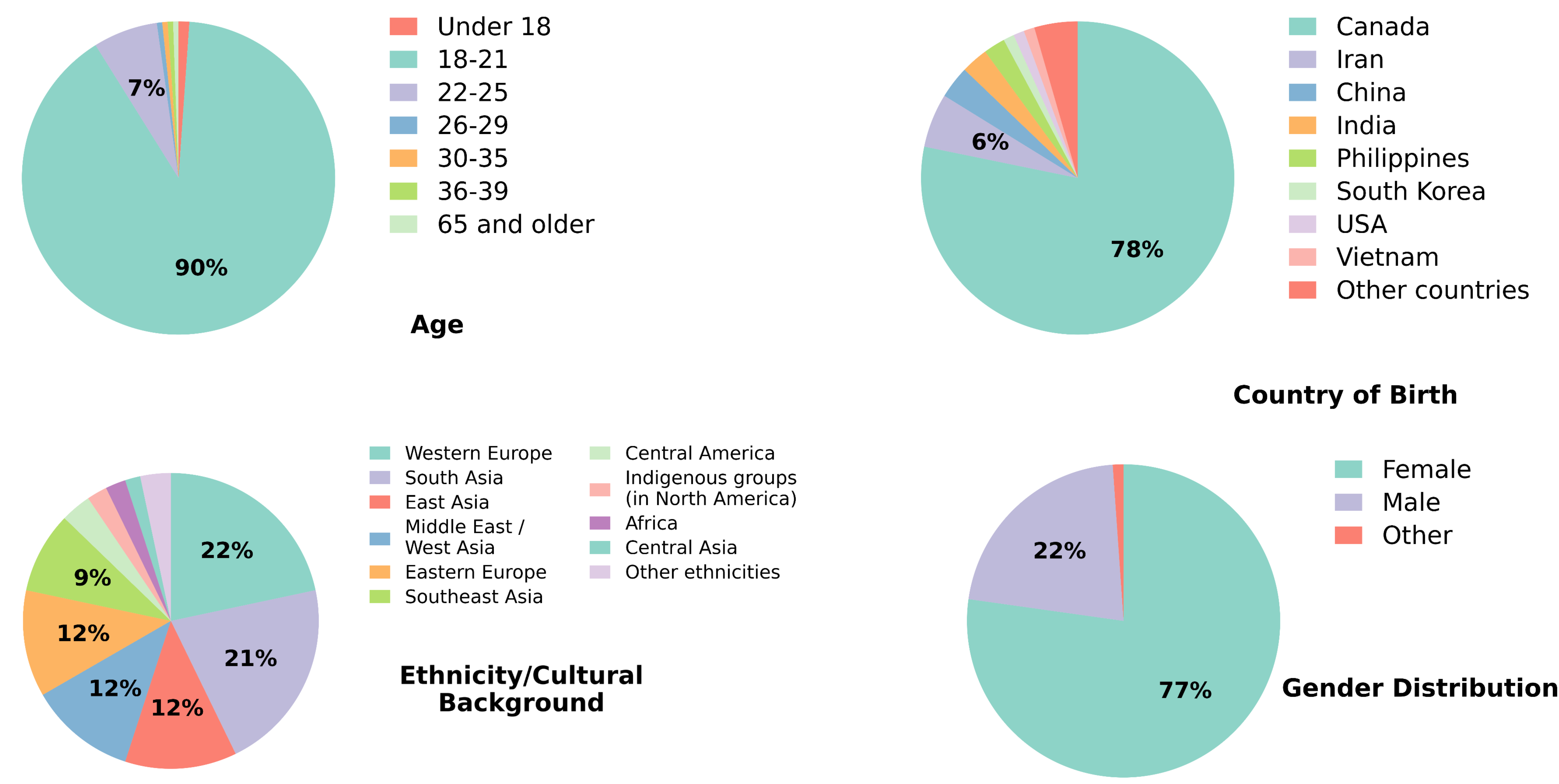}
    \caption{Demographic distribution of participants in the expression phase (Data Collection).}
    \label{fig:perceiver_dist}
\end{figure*}
\begin{figure*}
    \centering
    \includegraphics[width=\linewidth]{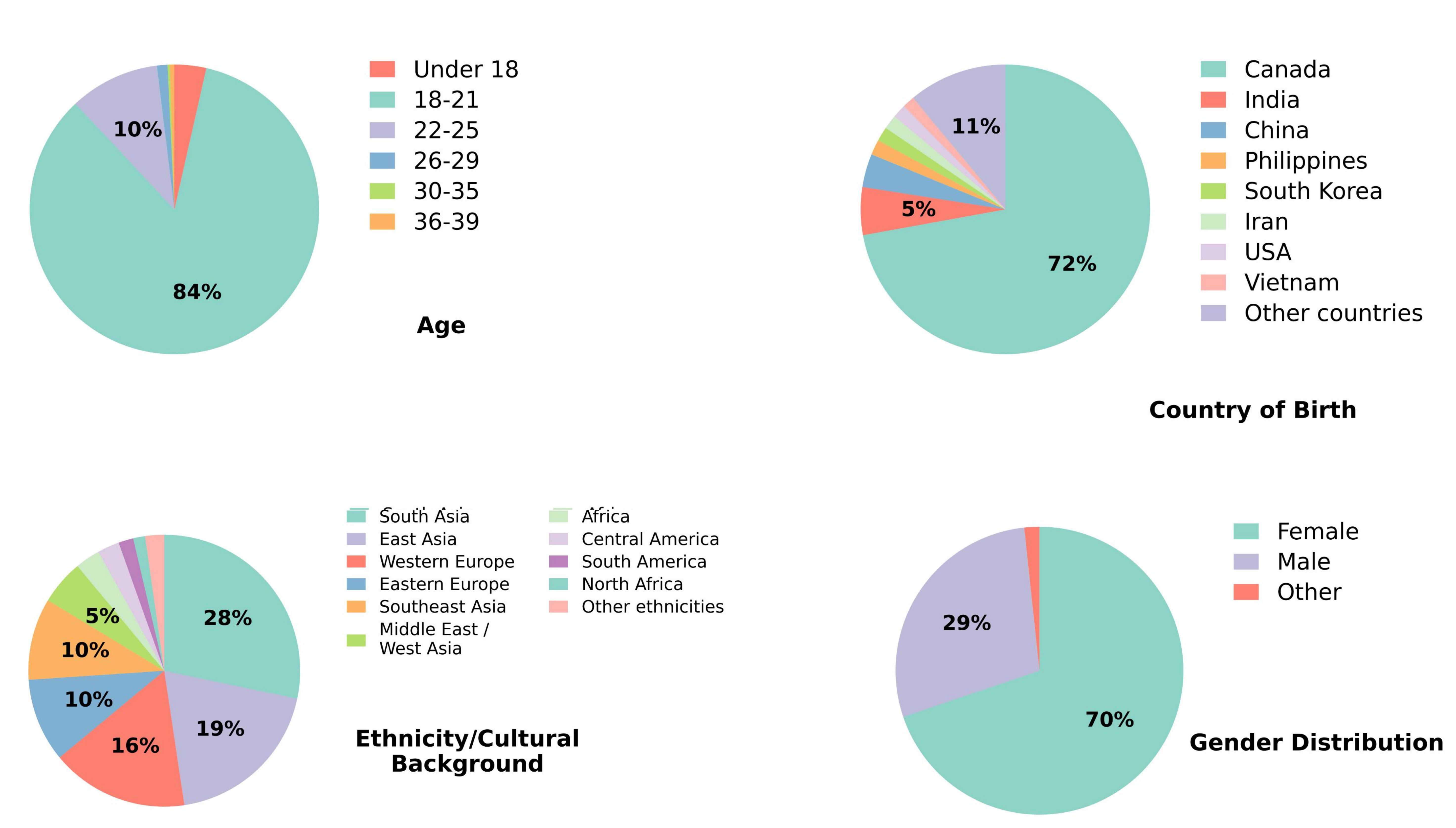}
    \caption{Demographic distribution of participants in the perception phase (Data Validation).}
    \label{fig:performer_dist}
\end{figure*}

\begin{table*}
\centering
\small
\begin{tabular}{p{0.22\textwidth} p{0.38\textwidth} p{0.32\textwidth}}
\hline
\textbf{Item} & \textbf{Question} & \textbf{Response options} \\
\hline
Age group 
& What is your age group? 
&  Under 18; 18-21; 22-25; 26-29; 30-35; 36-39; 40-49; 50-64; 65 and older\\

Gender identity 
& What is your gender identity? 
& Woman; Man; Other; Prefer not to say \\


Country of birth 
& Country of Birth 
& Country dropdown list\\

Ancestry or cultural identity 
& What is the region of origin or cultural identity of your ancestors? 
& North Africa; Africa excluding North Africa; Central America; South America; Eastern Europe; Western Europe; Middle East or West Asia; East Asia; Central Asia; South Asia; Southeast Asia; Polynesia or Pacific Islands; Indigenous groups in North America; Indigenous groups in Oceania; Other \\

Marital status 
& What is your current marital status? 
& Married; Common law; Divorced or separated; Widowed; Dating; Single \\
\hline
\end{tabular}
\caption{Demographic questionnaire completed by participants before the expression or perception task.}
\label{tab:demographic_questionnaire}
\end{table*}

The distributions of age, gender, country of birth, and ethnicity/cultural background for the human performers and annotators are shown in Figures~\ref{fig:perceiver_dist} and~\ref{fig:performer_dist}, respectively.

\section{Additional Experiment Details}
\label{appendix:benchmark_details}

\subsection{Metrics}
\label{appendix:metrics}
This section provides a more detailed explanation of the metrics, alongside their formulas. Let $v$ denote a video and $\ell$ denote a candidate expression label for that video. We use $\hat{\ell}_{v,k}$ for the model's $k$th top-ranked label, and $\ell^*_{v,k}$ for the human $k$th top-ranked label according to its mean rating across all annotators.

\textbf{Top-1}: We use ($T_1$) to check if the model's top prediction matches the top dominant human label.
\begin{equation}
    T_1(v) = \mathbb{1}[\hat{\ell}_{v,1} = \ell^*_{v,1}].
\end{equation}
We report the mean Top-1 score across all videos.

\textbf{Recall@2}: we use ($\mathrm{Recall_{@2}}$) to report how often the model recovers the two highest-rated labels, ignoring order.
\begin{equation}
\mathrm{Recall_{@2}}(v) =
    \frac{
    \left|
    \{\hat{\ell}_{v,1}, \hat{\ell}_{v,2}\}
    \cap
    \{\ell^*_{v,1}, \ell^*_{v,2}\}
    \right|
    }{2}
\end{equation}

\textbf{nDCG@2}: We use $nDCG_{@2}$ to report how often the model recovers the two highest-rated labels considering the order. The relevance score $r_v(\ell)$ is calculated using the top-1 and top-2 average human mean ratings of each video \cite{jarvelin2002cumulated}. 
For the model prediction $(\hat{\ell}_{v,1}, \hat{\ell}_{v,2})$, we compute:
\begin{equation}
    \mathrm{DCG_{@2}}(v) =
    r_v(\hat{\ell}_{v,1}) +
    \frac{r_v(\hat{\ell}_{v,2})}{\log_2 3}.
\end{equation}
The ideal DCG is obtained using the top two human labels:
\begin{equation}
    \mathrm{IDCG}_{@2}(v) =
    r_v(\ell^*_{v,1}) +
    \frac{r_v(\ell^*_{v,2})}{\log_2 3}.
\end{equation}
We then compute:
\begin{equation}
    \mathrm{nDCG}_{@2}(v) =
    \frac{\mathrm{DCG}_{@2}(v)}
         {\mathrm{IDCG}_{@2}(v)}.
\end{equation}

\textbf{Label Ratings MAE}: ($\textbf{MAE}$) measures the absolute difference between the computed mean human rating vector ($\mu^H_{v,j}$) and the computed mean model rating vector $\mu^M_{v,j}$ for candidate label $j$ of video $v$, where $D_v$ is the number of candidate labels:
\begin{equation}
    \mathrm{MAE}(v) =
    \frac{1}{D_v}
    \sum_{j=1}^{D_v}
    \left|
    \mu^H_{v,j} - \mu^M_{v,j}
    \right|.
\end{equation}
We report the Label Ratings MAE across videos.

\textbf{Rating-Distribution JSD}: We measure if the model can form the distribution of human Likert ratings for each candidate label.
Here, $X_{v,i,j}$ is the rating given by human annotator $i$ to label $j$ for video $v$, and $Y_{v,k,j}$ is the rating produced by the model in generation $k$ for the same video and label. We use $N_v$ and $K_v$ for the numbers of human annotators and model generations. For each rating score $s \in \{0,1,2,3\}$, we define the human and model rating distributions as:
\begin{equation}
    P^H_{v,j}(s) =
    \frac{1}{N_v}
    \sum_{i=1}^{N_v}
    \mathbb{1}[X_{v,i,j}=s],
\end{equation}
\begin{equation}
    P^M_{v,j}(s) =
    \frac{1}{K_v}
    \sum_{k=1}^{K_v}
    \mathbb{1}[Y_{v,k,j}=s].
\end{equation}
We then compute the Jensen-Shannon divergence between these two distributions for each label.
Let
\begin{equation}
    A_{v,j}(s) =
    \frac{1}{2}
    \left(
    P^H_{v,j}(s) + P^M_{v,j}(s)
    \right)
\end{equation}
be the average distribution. The label-level Jensen-Shannon divergence is:
\begin{equation}
    \begin{split}
        \mathrm{JSD}_{v,j} =
        \frac{1}{2}
        \sum_{s=0}^{3}
        P^H_{v,j}(s)
        \log
        \frac{P^H_{v,j}(s)}
        {A_{v,j}(s)}\\
        +
        \frac{1}{2}
        \sum_{s=0}^{3}
        P^M_{v,j}(s)
        \log
        \frac{P^M_{v,j}(s)}
        {A_{v,j}(s)}.
    \end{split}
\end{equation}

Then for each video rating JSD is computed by averaging over candidate labels, and we report the mean of rating JSD across all the videos.

\textbf{Coverage ($\mathrm{Cov.}$):} We use the Coverage metric described by Naeem et al.~\cite{naeem2020reliable} to measure what fraction of human rating vectors are represented by at least one model-generated rating vector. 
For each human rating vector $X_{v,i}$, let $r^{(k)}_{v,i}$ denote its Euclidean distance to the $k$-th closest rating vector among the other human rating vectors for the same video. A human response is covered if at least one model-generated rating vector is in the neighborhood: \begin{equation} 
C_{v,i} = \mathbb{1} \left[ \min_{1\leq m\leq K_v} \|X_{v,i}-Y_{v,m}\|_2 \leq r^{(k)}_{v,i} \right]. 
\end{equation} 
Coverage for video $v$ is then 
\begin{equation} 
\mathrm{Cov}(v) = \frac{1}{N_v} \sum_{i=1}^{N_v} C_{v,i}. 
\end{equation} 
We report the average coverage across videos. Higher values suggest that model-generated responses represent a larger proportion of the variation observed across human rating vectors. 

Using the hyperparameter-selection procedure proposed by Naeem et al.~\cite{naeem2020reliable}, we choose the smallest $k$ for which the expected Coverage exceeds $0.95$ when the human and model responses are drawn from the same distribution. The expected Coverage is \begin{equation}
\mathbb{E}[\mathrm{Cov}] =
1-\prod_{q=1}^{k}
\frac{N_v-q}{N_v+K_v-q}.
\end{equation}
where $N_v$ and $K_v$ are the numbers of human and model responses, respectively. With $\sim$30 human responses and $K_v=25$ model generations per video, this criterion gives $k=5$.

\textbf{Average Pairwise Distance Error (APDE):} We use Average Pairwise Distance Error (APDE)~\cite{barquero2023belfusion} to measure whether the model generates a similar response diversity as human annotators. Because the number of candidate labels varies across videos, we normalize the Euclidean distance between two rating vectors by the square root of the rating-vector dimensionality: 
\begin{equation} 
d_v(a,b) = \frac{\|a-b\|_2} {\sqrt{D_v}}. 
\end{equation} 
The human average pairwise distance for video $v$ is 
\begin{equation} 
\mathrm{APD}_H(v) = \frac{1}{\binom{N_v}{2}} \sum_{i<i'} d_v(X_{v,i},X_{v,i'}), 
\end{equation} 
and the corresponding model average pairwise distance is \begin{equation} 
\mathrm{APD}_M(v) = \frac{1}{\binom{K_v}{2}} \sum_{k<k'} d_v(Y_{v,k},Y_{v,k'}). 
\end{equation} 
APDE is the absolute difference between these two quantities: \begin{equation} 
\mathrm{APDE}(v) = \left| \mathrm{APD}_M(v) - \mathrm{APD}_H(v) \right|. 
\end{equation} 
We report the average APDE across videos. Lower APDE suggests that the amount of diversity among model-generated rating vectors is closer to that observed among human annotators.

\subsection{Implementation Details}
\label{appendix:imp_details}
To do our study, we used 4 NVIDIA L40S GPUs with 48GB memory. Models up to 8B parameters used 32 uniformly sampled video frames, while InternVL-38B used 24. Gemini-2.5-Flash~\cite{comanici2025gemini} (model ID: \texttt{gemini-2.5-flash}) was evaluated through the Google Gemini API, and GPT-5.6-Terra~\cite{openai2026gpt56} (model ID: \texttt{gpt-5.6-terra}, with reasoning effort set to \texttt{none}) was evaluated through the OpenAI Responses API. All API experiments were conducted in July and August 2026. Since GPT-5.6-Terra does not support native video input, the sampled frames were provided as temporally ordered images. Since \textit{Chehre} videos begin with a neutral expression, we removed 10\% of the frames from the beginning of the video before sampling. The average number of frames of the videos in our dataset is 67 frames. Therefore, after this trimming step, sampling 32 frames corresponds approximately to observing every two frames. This preprocessing standardizes the input across models and ensures that the sampled frames span most of each video.

\subsection{Prompts}
\label{appendix:prompts}
Here we provide the prompt used for model inference. For each video, the candidate labels were selected from the emoji-specific label set, shown in the prompt as \texttt{\{ALLOWED\_LABELS\}}, alongside the label \textit{None}. Here is the prompt used in the first benchmark task: dominant facial expression recognition.
\begin{tcolorbox}[
  breakable,
  colback=gray!15,
  colframe=gray!15,
  boxrule=0pt,
  arc=2pt
]
\begin{Verbatim}[fontsize=\small, breaklines=true, breakanywhere=true]
You are analyzing a short face video.
Task:
Classify the visible facial expression only.

Allowed categories:
{ALLOWED_LABELS}

Important:
- Return exactly two categories in the top_categories field.
- category one should be the most confident category.
- category two should be the second most confident category.
- if none of the categories are confident, return None for both category_1 and category_2.
- if only one category is confident, return None for the other category.
- Choose the categories from the allowed categories list only.

Required JSON format:
{{
"top_categories": [
    {"category_1": "CATEGORY_FROM_ALLOWED_LIST"},
    {"category_2": "CATEGORY_FROM_ALLOWED_LIST"}
],
"visual_cues": "short explanation based only on the face"
}}
\end{Verbatim}
\end{tcolorbox}
Here is the prompt used for the second benchmark task: distributional facial expression recognition. Compared with random seed sampling, persona prompting's prompt differed only in the observer-perspective sentence.
\begin{tcolorbox}[
  breakable,
  colback=gray!15,
  colframe=gray!15,
  boxrule=0pt,
  arc=2pt
]
\begin{Verbatim}[fontsize=\small, breaklines=true, breakanywhere=true]
You are an annotator analyzing a short face video.
{PERSONA_DESCRIPTION}
You are not describing the person's personality.
You are reporting how the facial expression appears to you as this annotator persona.

Task:
For each allowed facial expression label, assign an intensity score from 0 to 3.

Allowed categories:
{ALLOWED_LABELS}

Score each label from 0 to 3:
- 0 = No visible evidence of this expression.
- 1 = Weak or uncertain evidence of this expression.
- 2 = Clear evidence of this expression.
- 3 = Strong and obvious evidence of this expression.

Rules:
- Score every allowed label.
- Score each label independently.
- Do not rank the labels.
- Do not choose only the best label.
- Multiple labels can receive the same score.
- Use only integers: 0, 1, 2, or 3.
- Do not output null values in scores.
- Consider the persona when judging how the expression appears to you.
- The scores are from your perspective as the annotator persona.
- The persona must not invent facial evidence that is not visible.
- If the face clearly shows an expression, score it according to the visible evidence.
- The visual_cues field must describe only visible facial cues.
- Do not mention the persona in visual_cues.

Return strict JSON only:
{
  "persona": "{PERSONA_ID}",
  "scores": {
    "{LABEL_1}": null,
    "{LABEL_2}": null,
    "...": null
  },
  "visual_cues": "Brief explanation based only on visible facial cues."
}
\end{Verbatim}
\end{tcolorbox}
\textbf{The personas}:
Inspired by the Interpersonal Circumplex \cite{leary2004interpersonal}, we form personas based on affiliation and dominance:
The friendliness options are very unfriendly, unfriendly, neutral, friendly, very friendly.
The dominance status options are very submissive, submissive, balanced, dominant, very dominant.\\
For the prompt, we formed sentences based on these rules:
\begin{itemize}
\item No new sentence is added to the prompt if friendliness is \textit{neutral} and dominance status is \textit{balanced}.
\item\textit{Neutral} friendliness:
\texttt{You are someone who is [dominance status].}
\item\textit{Balanced} dominance status:
\texttt{You are someone who is [friendliness] toward people.}
\item Otherwise:
\texttt{You are someone who is [friendliness] toward people and is [dominance status].}
\end{itemize}

\newcommand{\emojirowfull}[4]{%
\includegraphics[height=1.5em]{images/emojis/#1.png} & \texttt{#1} & #2 & #3 \\}
\begin{table*}[t]
\centering
\scriptsize
\setlength{\tabcolsep}{2pt}
\renewcommand{\arraystretch}{1.05}
\begin{tabular}{c c c c c p{0.21\textwidth} | c c c c c p{0.21\textwidth}}
\hline
\textbf{Emoji} & \textbf{Code} & \textbf{Videos} & \textbf{High} & \textbf{Low} & \textbf{Labels}
& \textbf{Emoji} & \textbf{Code} & \textbf{Videos} & \textbf{High} & \textbf{Low} & \textbf{Labels} \\
\hline

\emojiimg{1f601} & \texttt{1f601} & 59 & 2 & 15 & Awkward, Excited, Happy, Overjoyed, Smiling
& \emojiimg{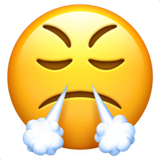} & \texttt{1f624} & 54 & 5 & 6 & Angry, Annoyed, Frustrated, Furious \\

\emojiimg{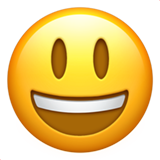} & \texttt{1f603} & 56 & 11 & 8 & Awkward, Excited, Happy, Sarcastic, Smiling
& \emojiimg{1f62b} & \texttt{1f62b} & 40 & 0 & 6 & Annoyed, Desperate, Distressed, In pain, Satisfied \\

\emojiimg{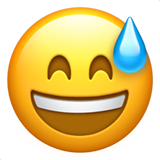} & \texttt{1f605} & 60 & 5 & 14 & Awkward, Embarrassed, Happy, Nervous, Relieved
& \emojiimg{1f62c} & \texttt{1f62c} & 63 & 0 & 19 & Awkward, Embarrassed, Nervous, Scared, Yikes \\

\emojiimg{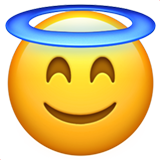} & \texttt{1f607} & 51 & 5 & 11 & Angelic, Happy, Innocent, Peaceful
& \emojiimg{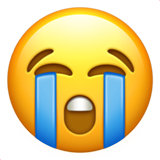} & \texttt{1f62d} & 29 & 4 & 4 & Crying, Devastated, Laughing, Sad \\

\emojiimg{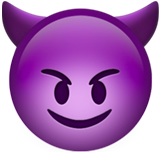} & \texttt{1f608} & 49 & 6 & 2 & Devious, Evil, Mischievous
& \emojiimg{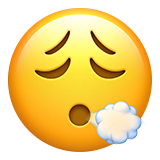} & \texttt{1f62e} & 40 & 3 & 9 & Exhausted, Impressed, Relieved, Smoking \\

\emojiimg{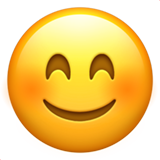} & \texttt{1f60a} & 58 & 6 & 2 & Blushing, Content, Happy
& \emojiimg{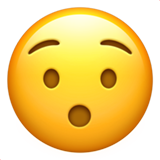} & \texttt{1f62f} & 60 & 7 & 2 & Amazed, Engaged, Shocked, Surprised \\

\emojiimg{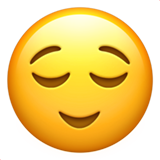} & \texttt{1f60c} & 52 & 19 & 1 & Content, Proud, Relaxed, Satisfied
& \emojiimg{1f630} & \texttt{1f630} & 53 & 18 & 1 & Anxious, Nervous, Scared, Stressed, Worried \\

\emojiimg{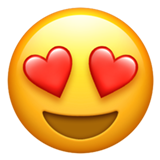} & \texttt{1f60d} & 48 & 11 & 3 & Admiring, Awestruck, In love
& \emojiimg{1f631} & \texttt{1f631} & 58 & 2 & 35 & Scared, Shocked, Surprised, Terrified \\

\emojiimg{1f60e} & \texttt{1f60e} & 48 & 20 & 2 & Confident, Cool, Smug
& \emojiimg{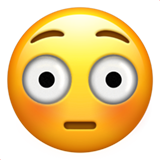} & \texttt{1f633} & 61 & 5 & 23 & Embarrassed, Flustered, Shocked, Shy, Surprised \\

\emojiimg{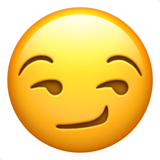} & \texttt{1f60f} & 60 & 8 & 1 & Flirty, Mischievous, Smirking, Smug
& \emojiimg{1f634} & \texttt{1f634} & 54 & 11 & 3 & Bored, Sleepy, Tired \\

\emojiimg{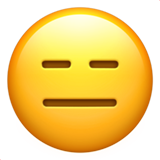} & \texttt{1f611} & 59 & 8 & 8 & Annoyed, Bored, Disappointed, Neutral, Unimpressed
& \emojiimg{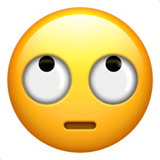} & \texttt{1f644} & 63 & 4 & 14 & Annoyed, Irritated, Rolling eyes, Sarcastic \\

\emojiimg{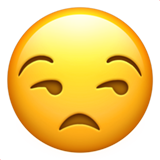} & \texttt{1f612} & 65 & 1 & 6 & Annoyed, Disgust, Judgmental, Side eye, Unimpressed
& \emojiimg{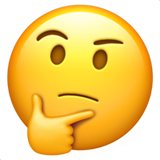} & \texttt{1f914} & 56 & 7 & 5 & Confused, Curious, Pondering, Thinking \\

\emojiimg{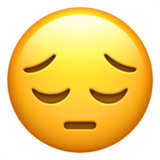} & \texttt{1f614} & 56 & 9 & 5 & Ashamed, Disappointed, Sad, Upset
& \emojiimg{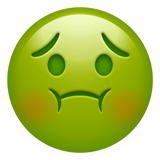} & \texttt{1f922} & 40 & 11 & 12 & Disgusted, Nauseous, Sick \\

\emojiimg{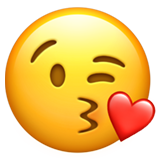} & \texttt{1f618} & 42 & 10 & 9 & Blowing a kiss, Flirty, Loving
& \emojiimg{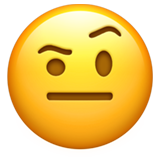} & \texttt{1f928} & 61 & 13 & 4 & Confused, Curious, Questioning, Suspicious \\

\emojiimg{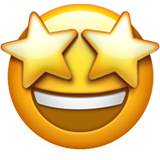} & \texttt{1f929} & 55 & 11 & 0 & Amazed, Awestruck, Excited, Surprised
& \emojiimg{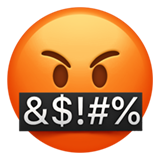} & \texttt{1f92c} & 52 & 17 & 2 & Angry, Enraged, Pissed off \\

\emojiimg{1f92d} & \texttt{1f92d} & 53 & 7 & 1 & Blushing, Cheeky, Embarrassed, Giddy, Giggly, Shy
& \emojiimg{1f92f} & \texttt{1f92f} & 54 & 3 & 9 & Mind blown, Shocked, Surprised \\

\emojiimg{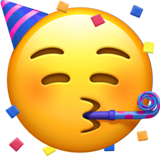} & \texttt{1f973} & 39 & 10 & 9 & Celebrating, Excited, Happy, Partying
& \emojiimg{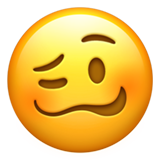} & \texttt{1f974} & 53 & 6 & 11 & Confused, Drunk, Silly, Weirded out \\

\emojiimg{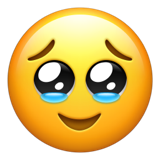} & \texttt{1f979} & 44 & 3 & 6 & Emotional, Grateful, Happy, Pleading, Proud
& \emojiimg{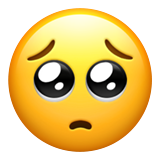} & \texttt{1f97a} & 45 & 6 & 5 & Begging, Cute, Pouting, Sad \\

\emojiimg{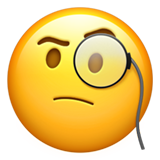} & \texttt{1f9d0} & 62 & 4 & 2 & Confused, Curious, Questioning, Skeptical, Thinking
& \emojiimg{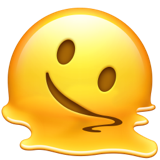} & \texttt{1fae0} & 45 & 9 & 3 & Defeated, Embarrassed, Melting, Overwhelmed, Smiling \\

\emojiimg{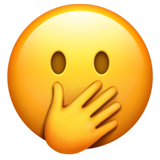} & \texttt{1fae2} & 53 & 2 & 16 & Embarrassed, Gasping, Shocked, Surprised
& \emojiimg{1fae4} & \texttt{1fae4} & 61 & 11 & 6 & Confused, Disappointed, Indifferent, Unsure, Upset \\

\hline
\end{tabular}

\caption{
Emojis with their final video counts, associated candidate label sets. High and Low indicate the number of videos for each emoji code in the high-ambiguity and low-ambiguity subsets, respectively.
}
\label{tab:emoji_label_counts}
\end{table*}
\section{Emojis and Labels}
\label{appendix:emoji_label_counts}
Table~\ref{tab:emoji_label_counts} shows the 40 emojis and the candidate label set used in our data, along with the number of videos for each emoji in the final dataset.

\section{Licensing}
Tools and models used in this work that are publicly available open-source projects for research purposes are listed here: LivePortrait~\cite{guo2024liveportrait}, StyleGAN2~\cite{karras2020analyzing}, EmoStyle~\cite{azari2024emostyle}, DeepFace~\cite{serengil2026boosted}, and MediaPipe~\cite{lugaresi2019mediapipe}. The Chicago Face Database~\cite{ma2015chicago} was used under its academic research license. The vision-language models evaluated in our benchmarks, Qwen2.5-Omni-7B~\cite{xu2025qwen25omnitechnicalreport}, LLaVA-Video-7B-Qwen2~\cite{zhang2025llavavideovideoinstructiontuning}, InternVL3-8B, and InternVL3-38B~\cite{zhu2025internvl3}, are all open-source and publicly available. 

\textit{Chehre} will be released under the Creative Commons Attribution-NonCommercial 4.0 International License (CC BY-NC 4.0). 

\end{document}